\documentclass[10pt,journal,compsoc]{IEEEtran}
\ifCLASSOPTIONcompsoc
  \usepackage[nocompress]{cite}
\else
  \usepackage{cite}
\fi

\ifCLASSINFOpdf
  \usepackage[colorlinks,
  linkcolor=black,
  anchorcolor=black,
  citecolor=black]{hyperref}  
  \usepackage{graphicx}
  \usepackage[caption=false,font=normalsize,labelfont=sf,textfont=sf]{subfig}
  \usepackage{amsmath,amsfonts}
  \usepackage{xcolor}
  \usepackage{tabularx}
  \usepackage{multirow}
  \usepackage{booktabs}
  \usepackage[linesnumbered,ruled]{algorithm2e}
  
  \SetCommentSty{mycommfont}
  
  \SetKwInput{KwInput}{Input}                
  \SetKwInput{KwOutput}{Output}              
\else
\fi

\begin{document}
%
\title{Differentiable Genetic Programming \\for High-dimensional Symbolic Regression}
%
%
%
%

\author{Peng Zeng,
        Xiaotian~Song,
        Andrew~Lensen,~\IEEEmembership{Member,~IEEE,}
        Yuwei~Ou,
        Yanan~Sun,~\IEEEmembership{Member,~IEEE,}
        Mengjie~Zhang,~\IEEEmembership{Fellow,~IEEE}
        and~Jiancheng~Lv,~\IEEEmembership{Senior~Member,~IEEE}
\IEEEcompsocitemizethanks{\IEEEcompsocthanksitem P. Zeng, X. Song, Y. Sun and J. Lv are with the College of Computer Science, Sichuan University, Chengdu 610064, China.\protect\\
E-mails: \{zengpeng7,songxt,ouyuwei\}@stu.scu.edu.cn,\\\{ysun,lvjiancheng\}@scu.edu.cn
\IEEEcompsocthanksitem  A. Lensen and M. Zhang are with the Evolutionary Computation Research Group, Victoria University of Wellington, Wellington 6140, New Zealand.\protect\\
E-mails:\{andrew.lensen,mengjie.zhang\}@ecs.vuw.ac.nz
}
}

\IEEEtitleabstractindextext{%
\begin{abstract}
Symbolic regression (SR) is the process of discovering hidden relationships from data with mathematical expressions, which is typically considered an effective way to reach interpretable machine learning. Genetic programming (GP) has been the dominator in solving various SR problems. However, as the scale of SR problems increases, GP often poorly demonstrates and cannot effectively address the real-world high-dimensional problems. This limitation is mainly caused by the stochastic evolutionary nature of traditional GP in constructing the proper trees. In this paper, we propose a differentiable approach named DGP to construct GP trees towards high-dimensional SR for the first time. Specifically, a new data structure called differentiable symbolic tree is proposed to relax the discrete structure to be continuous, thus a gradient-based optimizer can be easily presented for the efficient optimization. In addition, a sampling method is proposed to eliminate the discrepancy caused by the above relaxation for valid symbolic expressions. Furthermore, a diversification mechanism is introduced to promote the optimizer escaping from local optima for globally optimal expressions. With these designs, the proposed DGP method can efficiently search for the GP trees with higher performance, thus being capable of dealing with high-dimensional SR. To demonstrate the effectiveness of DGP, we conducted various experiments against the state of the arts based on both GP and deep neural networks. The experiment results reveal that DGP can outperform these chosen peer competitors on real-world high-dimensional regression benchmarks with dimensions varying from hundreds to thousands. In addition, on the challenging synthetic SR problems, the proposed DGP method can also achieve the best recovery rate even with different noisy levels. It is believed this work can facilitate SR being a powerful alternative to interpretable machine learning for a broader range of real-world problems.
\end{abstract}

\begin{IEEEkeywords}
Symbolic regression, genetic programming, gradient descent, interpretable machine learning, neural networks.
\end{IEEEkeywords}}

\maketitle

\IEEEdisplaynontitleabstractindextext
%
\IEEEpeerreviewmaketitle

\IEEEraisesectionheading{\section{Introduction}\label{sec:introduction}}

%
%
%
%

\IEEEPARstart{I}{n} scientific research, a key task is to discover relationships within experimental data and then formalize them into symbolic expressions~\cite{Schmidt.2009}. This process is widely performed using symbolic regression (SR)~\cite{Awange.2016}, a fundamental machine learning technique. Research into SR has been steadily ongoing for decades~\cite{Langley.1981}, with research interest largely based on the inherent interpretable nature of symbolic expressions. The recent growth in interpretable artificial intelligence has placed SR as a rapidly-emerging topic in recent years~\cite{Udrescu.2020,Virgolin.2020,Petersen.2021,Biggio.2021,Chen.2022}.

In the literature, there are two mainstream approaches to achieving SR: neural networks (NNs)~\cite{A.Prieto.2016} and genetic programming (GP)~\cite{Koza.1994}. The NN-based SR approaches can be further classified into two different categories based on whether the symbolic expressions are constructed directly or indirectly. Symbolic expressions in the direct NN-based approaches are generated from the NNs without any intermediate agent. These approaches first construct an NN in the same way as in traditional regression tasks, and replace the traditional activation functions with mathematical symbols. The constructed NN is trained with sparse constraints~\cite{hosseini2015deep} and then the mathematical symbols associated with significant weights are sequentially extracted to create the symbolic expressions. The canonical work by Martius \textit{et al.}~\cite{Martius.2016} constructed a fully-connected NN with three symbols, i.e., $=$, $sin(\cdot)$, and $cos(\cdot)$. Sahoo \textit{et al.}~\cite{Sahoo.2018} later extended this work adding the ``$\div$'' operator to allow for application to a wider variety of problems. Kim \textit{et al.}~\cite{Kim.2021} posited that a fully-connected NN cannot effectively address sequential SR problems; they proposed using multiple NNs to perform SR in kinematics, where each NN is trained on a specific time slot. Other types of NNs have also been used: Cranmer \textit{et al.}~\cite{Cranmer.2020} utilized a graph NN~\cite{scarselli2008graph} to simulate physical particle systems. They suppose that data in the physical particle systems are often non-Euclidean, which makes the graph NN an ideal model for such kinds of data.

In contrast, indirect NN-based SR approaches utilize intermediary agents to generate symbolic expressions. These agents are often trees composed of mathematical symbols. Such approaches view SR as a special structure-generation task, using generative models based on NNs (such as recurrent NNs (RNNs)~\cite{Ayyadevara.2018}) to progressively generate a tree structure. After that, the trees are transformed into the final symbolic expressions. The motivation for this approach is that the generated expressions do not directly depend on the structure of NNs: thus complex and efficient NNs can be utilized for more challenging SR problems. There are a few representative works in this category~\cite{Petersen.2021,Kusner.2017,Mundhenk.2021}, which differ in how they select different generative models according to different goals, such as accuracy or speed. For instance, Kusner \textit{et al.}~\cite{Kusner.2017} proposed a special variational encoder-decoder~\cite{Kingma.2013} structure with grammar rules to map the trees representing symbolic expressions into a latent space and then generate valid expressions from it. The encoder-decoder structure has been shown to be effective and was also used in SymFormer~\cite{MartinVastl.2022}, which is based on the  Transformer architecture~\cite{Vaswani.2017}. Petersen \textit{et al.}~\cite{Petersen.2021} and Mundhenk \textit{et al.}~\cite{Mundhenk.2021} formulated the SR problem as a reinforcement learning problem and used an RNN to generate expression trees over multiple iterations to search for more accurate symbolic expressions.

Although the above NN-based SR approaches have shown effectiveness in experiments, limitations remain. Firstly, the performance of an NN highly depends on its structure, which is often manually designed, requiring expertise in both NNs and in the domain of the problem being solved~\cite{Biggio.2021}. However, such kind of knowledge is often not held by users~\cite{Sun.2020}. Secondly, due to the non-convexity of NNs, these approaches easily get stuck in local optima, resulting in poorly performing, inexact symbolic expressions. This phenomenon has been reported in both the direct and the indirect NN-based SR approaches in very recent literature~\cite{HaoranLi.2022}. 

GP, an evolutionary learning method that automatically constructs tree structures~\cite{Koza.1994}, has the advantages of not requiring prior knowledge and also being robust to local minima~\cite{Langdon.2008}. Thus, GP is naturally used for SR problems. In canonical GP-based SR, symbolic expressions are represented as trees, where in each tree, the leaf nodes are variables/features, and the non-leaf nodes are operators. GP maintains a population of multiple trees (which have different structures) and evolves the population generation-to-generation by using genetic operators such as crossover and mutation to change the structure as well as parameters. A pre-defined fitness function is used to evaluate each tree in the population. Based on their fitness, trees are selected as parents by a selection operator for generating individuals in the next generation. The evaluation and selection operators guide the GP search towards the optimal symbolic expression. In recent years, various GP methods have been designed for SR problems~\cite{Uy.2011,Lu.2016,Martins.2018,Virgolin.2020,Chen.2022}. Unfortunately, these methods are mainly designed for low-dimensional problems and cannot effectively scale well to high-dimensional problems\footnote{We note that there is no formal definition of ``high dimensional'' in the literature. Based on recent work~\cite{Udrescu.2020,Virgolin.2020,Petersen.2021,Biggio.2021,Chen.2022} and also our own knowledge of challenging SR problems, we consider an SR problem to be high-dimensional when its input has a dimensionality over 50.} that are prevalent in our big data era~\cite{gpsurvey2018}. 

While the flexible representation of GP makes it good at evolving diverse forms of symbolic expressions, when facing high-dimensional input variables, it tends to generate overly complex models that overfit the training data~\cite{Chen.2022}. To address this problem, multiple works have been proposed. One popular category of methods utilizes geometric semantic GP (GSGP)~\cite{Vanneschi.2017} in SR to improve the generalization performance~\cite{Pawlak.2015,Martins.2018,Chen.2019}. GSGP makes use of semantic-aware operators to enhance the variation process of GP by considering intermediate solution outputs (i.e. partial solutions within a tree), making the search process more efficient. The method enables GP to find more accurate expressions, but often at the cost of complexity~\cite{MarcoVirgolin.2019}. Furthermore, the expression size of GSGP-based SR varies from hundreds to millions of components~\cite{Pawlak.2015}\cite{Martins.2018}, which makes it almost impossible for humans to understand these expressions. Researchers have also explored combining GP and feature engineering to improve the generalization performance of GP on high-dimensional problems~\cite{Chen.2017,ChenQiandZhangMengjieandXueBing.2017}. For instance, Chen et al.~\cite{Chen.2017} try to improve the generalization performance of GP for high-dimensional SR by using feature selection. They propose a new feature selection method based on permutation and use it to select the truly relevant features for GP-based SR. The new improvement experimentally alleviates the generalization performance degradation of GP on high-dimensional SR tasks. However, the inefficiency of GP evolutionary search approach high-dimensional problems, which causes the above problem from the root, remains to be addressed. 

Another issue is that the interpretability of GP solutions suffers substantially on high-dimensional problems. GP often leads to very large expressions or chaining of nonlinear functions, which makes the expressions much harder to analyze~\cite{Vanneschi.2010}. This problem is related to ``bloating'' in GP (where trees grow unnecessarily large without a corresponding increase in fitness), which is a common side effect of the evolutionary-based search approach in GP. The flexible representation of GP means it can theoretically discover the entire set of possible function forms. However, this enormous search space includes redundant expressions; the evolutionary process often generates unnecessarily large models with poor interpretability~\cite{Franca.2021}. Bloating is a common problem in GP-based SR and has been researched for decades. Previous solutions to this problem include incorporating a parsimony measure to the fitness function~\cite{Vladislavleva.2009,FabricioOlivettideFranca.2018} or utilising online program simplification~\cite{Kinzett.2009}. For instance, Kinzett \textit{et al.}~\cite{Kinzett.2009} explored the use of two online program simplification approaches in the evolution process: algebraic simplification~\cite{Wong.2006} and numerical simplification. This helps to reduce bloating in the evolution of GP. However, it needs predefined parameters, i.e., simplification rules and threshold. The final test performance is highly sensitive to these parameters, and it is hard to preset them properly. More recently, Franca \textit{et al.}~\cite{FabricioOlivettideFranca.2018} introduced a new representation called interaction-transformation, which constrains the search space in order to exclude a region of larger and more complicated expressions. The approach has proven effective in traditional small-scale SR problems, but it does not scale well as problem dimensionality increases~\cite{Franca.2021}. 

In summary, GP-based SR approaches have the potential to be interpretable. However, the stochastic evolutionary-based search makes finding the optimal structure inefficient, resulting in performance degradation on high-dimensional problems. In contrast, the NN-based SR approaches have high search efficiency due to their gradient-based optimization way - but the black-box nature of NNs makes it hard to convert an NN into an interpretable expression. To combine the benefits of both GP and NNs, we for the first time propose \textbf{differentiable GP (DGP)}, which is a gradient-based algorithm to effectively construct GP trees on high-dimensional SR problems. The main contributions of the proposed DGP algorithm are listed as follows:
\begin{itemize}
	\item We propose a representation, the differentiable symbolic tree, in DGP that relaxes the discrete tree structure into a continuous space. This design allows the structure to be optimized using a gradient-based optimizer. As a result, DGP can more efficiently search for symbolic expressions with higher performance on high-dimensional problems.
	
	\item We propose a sampling method that can map changes in continuous value to discrete structures, allowing the creation of valid symbolic trees. Furthermore, we introduce a diversification mechanism that acts on the symbolic trees to produce more diverse structures. This design improves the global search ability of DGP, allowing the exploration of symbolic expressions across the whole search space.
	
	\item We show that the proposed DGP method can outperform existing SR methods (based on both GP and deep NNs) on various high-dimensional, real-world SR benchmarks. Furthermore, we also demonstrate that DGP can recover the exact symbolic expressions on the synthetic datasets across different levels of noise. We conclude that DGP is an effective tool for challenging, real-world SR problems.
\end{itemize}

The remainder of this article is structured as follows. First, background and related works are presented in Section~\ref{background}. Then, the proposed DGP method is described in detail in Section~\ref{method section}. Next, the experiment design and experiment results, as well as the analysis (including recovery and robustness analyses), are given in Sections~\ref{experiment} and \ref{result}. Finally, we conclude the paper and present our vision for future work in Section~\ref{conclusion}.

\section{Background and Related Work}\label{background}
In this section, we introduce crucial background knowledge in SR. We then review the relevant work that uses GP with gradient descent to demonstrate the novelty of our proposed DGP method. 

\subsection{Symbolic Regression (SR)}
SR is a sub-field of machine learning concerned with discovering an analytical model of the given data in the form of a symbolic expression or mathematical formula. In general, the training process of SR can be formalized as Equation~(\ref{eq1}):
\begin{equation}
	f^{*} = arg min_{f}\sum_{i=1}^{n}D(y_{i}, f(X_{i}))\label{eq1}
\end{equation}
where $X_{i} \in \mathbb{R}^{d}$ and $y_{i} \in \mathbb{R}$ are the input-output pairs with continuous values in a dataset with size $n$. $f(\cdot):\mathbb{R}^{d}\rightarrow\mathbb{R}$ represents a function in the form of a symbolic expression and $D(\cdot,\cdot)$ is a distance function. The goal of SR is to find a symbolic expression $f^{*}$ that minimize $D(y_{i}, f(X_{i}))$ for $\{(X_{i}, y_{i})| i=1,2,3,\cdots, n\}$. After training, the expression $f^{*}$ can be used to analyze the underlying relationships within the data and can also be used as a model to make predictions on unseen data. 

SR has two significant differences from other regression techniques. Firstly, compared to statistical regression techniques, which only perform parameter optimization on a pre-defined fixed model structure, SR can adaptively discover a suitable model structure with appropriate parameters. For example, linear regression assumes a linear relationship between the input variables and the output by using linear equations, where the linear coefficients need to be optimized. However, SR does not pre-determine the structure of the expression, and the expressions obtained can be linear, polynomial, or of other shapes. This means that SR is a mixed optimization process, where the optimal structure and the parameters are searched at the same time. This advantage allows SR to have broader applications than statistical regression techniques, especially in the face of unknown problems, where domain experience is often lacking, as would be needed to pre-design a model structure.

Secondly, other numerical regression methods, such as NNs, are generally treated as black-box models and only judged by their prediction accuracy on unseen data. SR, while producing accurate results, is also particularly well suited for human interpretability and in-depth analysis, as the expressions identify the true underlying functional relationship behind the data. For instance, given a dataset consisting of $x \in \mathbb{R}$ and $y \in \mathbb{R}$, if we use an NN for regression, we only get a black-box NN model for prediction after training. Instead, if we use SR to process it, we can obtain a symbolic expression, such as $y = -9.8 \times sin(x)$. We can not only make predictions but also reflect that $x$ and $y$ might be the angle and angular velocity taken from the same pendulum system, based on which performing further analysis. Given such advantages, SR is considered to be an effective way to enable interpretable machine learning~\cite{Ferreira.2020}.

\subsection{Genetic Programming (GP) with Gradient Descent}
GP is famous for its ability to automatically generate Lisp-style computer programs and was initially popularised by Koza~\cite{Koza.1994}. GP can easily represent computer programs using hierarchical syntax trees, which means GP is suitable for automatic programming. With such tree structures, the size, shape, and contents of the program can be changed dynamically by genetic operators of GP. In SR, the symbolic solutions are similar to automatic programming, so GP was naturally extended to SR and gradually became the dominant technique among various SR techniques~\cite{PatrykOrzechowski.2018}. 

The idea of combining GP with gradient descent search to enhance GP for SR has been researched before. The initial works in this area were FGP~\cite{Topchy.2001} and WLGP~\cite{Zhang.2005}. FGP utilized a gradient descent search to optimize the numeric leaf values for GP and found that the gradient search can provide improvements in accuracy as a form of local search. In WLGP, inclusion factors are introduced to different subtrees of GP and the gradient descent search is utilized to optimize them, which collectively increases the overall performance. The gradient descent approach in both FGP and WLGP was applied to every program in the population, which negatively impacts the efficiency of the search process~\cite{Graff.2014}--performing gradient descent search for all the programs in the population comes at additional computation cost. Later, Chen \textit{et al.}~\cite{Chen.2015} focused on balancing accuracy and efficiency by applying gradient descent to only the top 20\% programs in the population. In recent work, Dick \textit{et al.}~\cite{Dick.2020} found that feature standardization can improve the efficiency of gradient descent: they proposed a hybrid method that combined feature standardization and stochastic gradient descent. 

In summary, these existing methods only use gradient descent search as a complement to evolutionary methods for local search. However, in practice, the gradient descent approach has already been proved to be effective in both parameter and structure optimization in the filed of NN, therefore it is also worth exploring to utilize this efficient approach to optimize the structure of GP. Unfortunately, the gradient descent approach has never been used to dominate the optimization of the GP structure. This is because the optimization of GP structure is a fundamentally discrete problem. The first contribution of this study is just to address this critical issue.

\section{The Proposed Method}\label{method section}
In this section, we describe the proposed DGP method in detail. First, we present the overall framework of DGP in Subsection~\ref{OverallFramework}. Then, we detail the optimization and diversification process, which are the key steps in DGP, in Subsections~\ref{optimization} and \ref{diversification}, respectively.

\subsection{Overall Framework} \label{OverallFramework}
The overall framework of DGP is described in Algorithm~\ref{algorithm1}. By giving the pre-defined number of iterations $G$ and the training dataset $D$, DGP first initializes the expressions (line~\ref{line_1}), and then performs a series of iterations (lines~\ref{line_1_1}-\ref{line6_6}). Finally, the best searched expression regarding $D$ is obtained (line~\ref{line_7}).
\begin{algorithm}[htp]
	\DontPrintSemicolon
	
	\KwInput{The maximal iterations $G$, the dataset $D$}
	\KwOutput{The optimal expression}
	
	$\tau_0$ $\leftarrow$ Initialize expressions\; \label{line_1}
	\For{t = 1, ..., G}{\label{line_1_1}
		$\tau \leftarrow \emptyset$ \; \label{line_2}
		\For{$e$ in $\tau_{t-1}$}{      \label{line_3}
			$\tau \leftarrow \tau \cup$ \textbf{Optimization}($e$, $D$) \label{line_4}
		}\label{line_4_1}
		\textbf{Diversification}($\tau$, $D$)\; \label{line_5}
		
		$\tau_t$ $\leftarrow$ $\tau$\; \label{line_6}
	}\label{line6_6}
	\textbf{Return} the best solution in $\tau_t$\; \label{line_7}
	\caption{The Overall Framework of DGP}
	\label{algorithm1}
\end{algorithm}

In the initialization process, an expression set $\tau_0$, consisting of mathematical expressions, is initialized. Please note that there is no specific prior knowledge introduced to this initialization -- $\tau_0$ contains only the most basic mathematical expressions used by other GP tasks, such as $x$, $x+y$, etc. During each iteration $t$, an empty set $\tau$ is created to store the expressions found in that iteration (line~\ref{line_2}). After that, the process of \textit{Optimization}, i.e., the gradient-based optimization designed in DGP, is performed on $D$ for each mathematical expression $e$ in $\tau_{t-1}$, and then the optimized $e$ is added to $\tau$ (lines~\ref{line_3}-\ref{line_4_1}). Next, the process of \textit{Diversification} is performed on the optimized expression set $\tau$ (line~\ref{line_5}).

As shown above, the search process of DGP consists of two important parts: \textit{Optimization} and \textit{Diversification}, which are described in the following subsections.

\subsection{Optimization}\label{optimization}
\begin{figure}[h] 
	\centering 
	\includegraphics[width=0.5\textwidth]{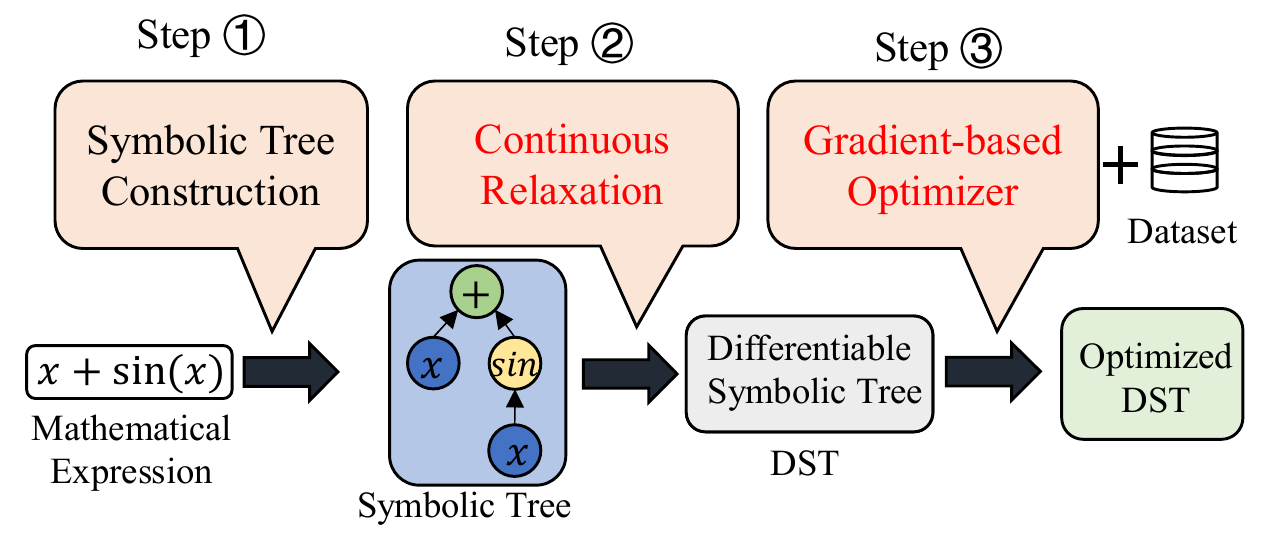}
	\caption{Illustration of the optimization process. The optimization process can be divided into three steps. Step 1: Given an expression, a symbolic tree is constructed to represent it. Step 2: The symbolic tree is relaxed to a continuous model called differentiable symbolic tree (DST). Step 3: The DST is evaluated on the training dataset to compute loss and then optimized by gradient descent.} 
	\label{gradient_descent}
\end{figure}

The optimization phase is proposed to search for the best structure in a subspace efficiently. The overall process of optimization can be divided into three steps. They are: \textbf{Symbolic Tree Construction}, \textbf{Continuous Relaxation}, and the \textbf{Gradient-Based Optimizer}, which corresponds to Step \textcircled{1} to \textcircled{3} in Fig. \ref{gradient_descent}. In Step \textcircled{1}, the mathematical expression is represented by the symbolic tree. In Step \textcircled{2}, the symbolic tree is relaxed into a continuous model called a differentiable symbolic tree (DST). In Step \textcircled{3}, the DST is optimized with gradients iteratively, and the optimized DST is finally obtained.

Among the three steps, the first step follows the same process as other GP-based SR approaches. The function set used is \{$+$, $-$, $\times$, $\div$, $sin$, $cos$, $exp$, $log$\} and the terminal set is the input variables in the datasets, which together make up the primitive set. In the following subsections, we will focus on the last two steps.

\subsubsection{Continuous Relaxation} \label{subsec:continous relaxtion}
In traditional GP-based SR, the search space of model structures is discrete. Specifically, a primitive in one node of a symbolic tree can randomly change to another through genetic operators of mutation and crossover in the evolutionary process. In DGP, we propose optimizing the structure of the symbolic trees by leveraging the gradient information, which requires a continuous search space. Therefore, \textit{continuous relaxation} is proposed to construct a continuous subspace from a given tree structure. We achieve this through the use of a node matrix and adjacency matrix, which will be introduced below. 

\begin{figure*}[h] 
	\centering 
	\includegraphics[width=\textwidth]{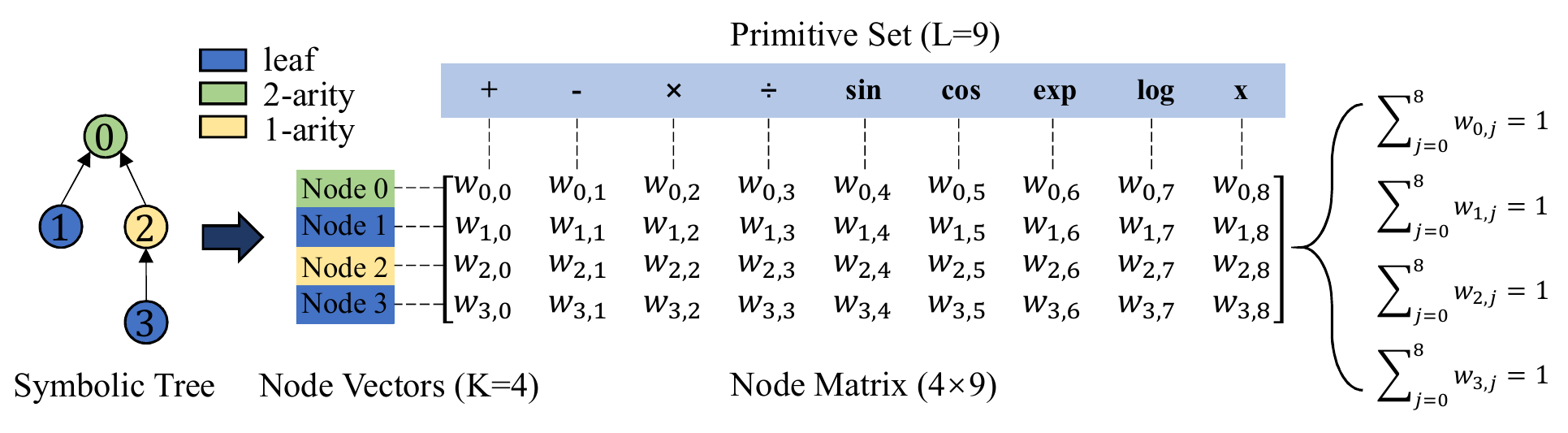}
	\caption{Illustration of the generation of node matrix. For each node in the symbolic tree, a vector having a length equal to the size of the primitive set is initialized with zeros, and then the element is set to one if it is contained by the primitive set. These vectors together form the node matrix. During the GP structure optimization, the values of the elements in the node matrix are changed with the constraint that the sum of each vector is equal to one.} 
	\label{node_mat_pic}
\end{figure*}

\textbf{Node Matrix}: assuming that the total number of nodes in the tree is $K$, the node matrix is composed of $K$ vectors. The dimension of each vector is $L$, which is the length of the primitive set. The values in the vector are floats that represent the proportion of the operation or variable in the node, which can also be viewed as the probability of the node taking
a certain operation or variable. As a result, the node matrix can be formulated by Definition~\ref{definition1}.

\newtheorem{myDef}{Definition}
\begin{myDef}\label{definition1}
	\textbf{Node matrix N} is a $K\times L$ matrix that satisfies Equation~(\ref{Node_matrix}):
	\begin{equation}
		\left\{
		\begin{aligned}
			N &=\binom{N_{nonleaf}}{N_{leaf}}\in\mathbb{R}_{+}^{K \times L} \\
			&=\begin{pmatrix}
				w_{0,0} & ... & w_{0,L-1} \\
				...& w_{i,j} & ... \\
				w_{K-1,0}&...  & w_{K-1,L-1} \\
			\end{pmatrix} \\
			s.t. &\sum_{j=0}^{L-1}w_{i,j} =1, i = 0, ..., K-1.
		\label{Node_matrix}
		\end{aligned}\right.
	\end{equation}
\end{myDef} 
where $N_{nonleaf}\in R_{+}^{n\times L}$ and $N_{leaf}\in R_{+}^{m\times L} $ are for non-leaf nodes and leaf nodes of the symbolic tree, respectively. $n$ and $m$ are the numbers of non-leaf nodes and leaf nodes in the tree. Let $\hat{w_{ij}}$ be the original value in $N$. A softmax activation is applied to derive $w_{ij}$, i.e., $w_{ij} = {exp(\hat{w_{ij}})}/{\sum_{j^{'}=1}^{L}exp(\hat{w_{i,j^{'}}})}$, to implicitly encode the constraint $\sum_{j=1}^{L}w_{ij} =1$. This constraint guarantee each row of $N$ is a valid primitives' distribution. $\hat{w}$ and the corresponding $w$ are set to be learnable, which guarantees that the proportion of different primitives in each node in the tree can be changed by training. An example of the generation of $N$ is illustrated in Fig.~\ref{node_mat_pic}. As is shown in the figure, the symbolic tree has 4 nodes, numbered 0 to 3. Each node can be represented by a vector comprising all the primitives. Vectors 0 to 3 are used to represent nodes 0 to 3 correspondingly. In the example, the length of the primitive set is 9, so each vector contains 9 weights representing the proportion of each primitive, giving a 4$\times$9 node matrix.

\begin{figure}[h] 
	\centering 
	\includegraphics[width=0.4\textwidth]{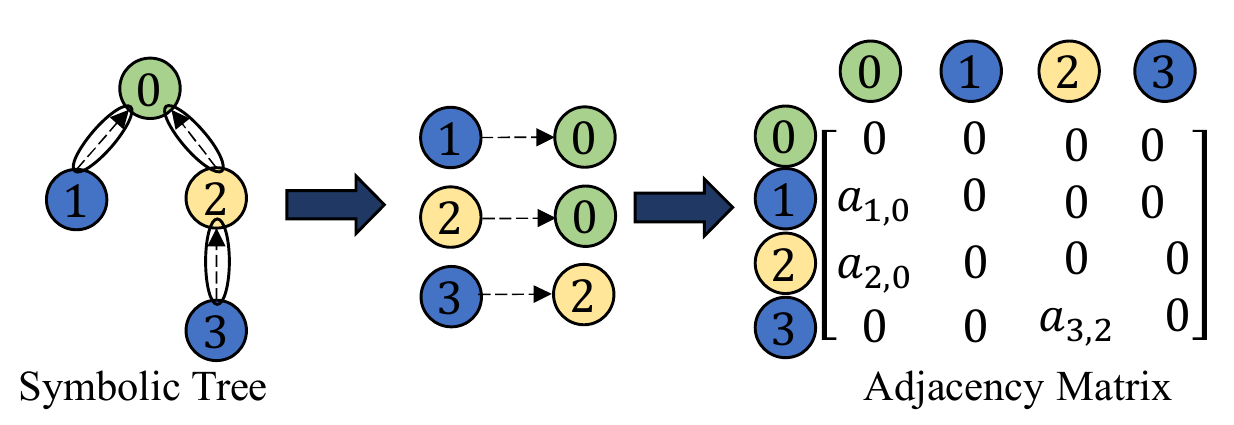}
	\caption{Illustration of the adjacency matrix. $a_{i, j}$ represents that node $i$ is the child of node $j$, and $0$ means that there is no connection between the two nodes.} 
	\label{adj_mat_pic}
\end{figure}

\textbf{Adjacency Matrix}: given the total number of nodes in the tree is $K$, the adjacency matrix is of shape $K\times K$. Each row and column in the matrix represent the adjacency information of a node. Specifically, each row in the matrix represents which nodes that the node points to, i.e., the parent node of it, and each column represents the child nodes of the current node. The values in the adjacency matrix indicate the strength of the connection between the nodes, which will be used when a node is deleted or replaced in the sampling step. Formally, the adjacency matrix is defined by the Definition~\ref{definition2}.

\begin{myDef}\label{definition2}
	\textbf{Adjacency matrix A} is a $K \times K$ matrix that satisfies Equation~(\ref{ADJ_matrix}):
	\begin{equation}
		\left\{
		\begin{aligned}
			A\in\mathbb{R}^{K \times K} = 
			\begin{pmatrix}
				a_{11} & ... & a_{1K} \\
				...& a_{ij} & ... \\
				a_{K1}&...  & a_{KK} \\
			\end{pmatrix}, \\
			where \ a_{ij} = \left\{\begin{matrix}
				\sigma\left ( v_{ij} \right ) & , connected \\
				0& ,disconnected \\
			\end{matrix}\right. \label{ADJ_matrix}
		\end{aligned}\right.
	\end{equation}
\end{myDef}
where $v\in\mathbb{R}^{K}$ are the parameters, and the sigmoid function $\sigma \left (\cdot \right )$ imposes the constraint $0\leq a_{ij}\leq1$. For connected nodes $i$ and $j$, $a_{ij} = \sigma (v_{ij})$ measures the strength of the connection. The illustration of the generation process of A is given in Fig.~\ref{adj_mat_pic}. As is shown in the figure, the symbolic tree has 4 nodes, numbered 0 to 3, thus its adjacency matrix is $4\times 4$. The values in the matrix are initialized according to the connections in the tree. Specifically, there are 3 directed edges in the tree. They are node 1$\rightarrow$node 0, node 2$\rightarrow$node 0, and node 3$\rightarrow$node 2. So there are 3 values greater than 0 in the matrix, which are $a_{1,0}$, $a_{2,0}$, and $a_{3,2}$, respectively. The rest of the values are 0, which means that there is no connection between the corresponding nodes.

\begin{figure*}[h] 
	\centering 
	\includegraphics[width=0.85\textwidth]{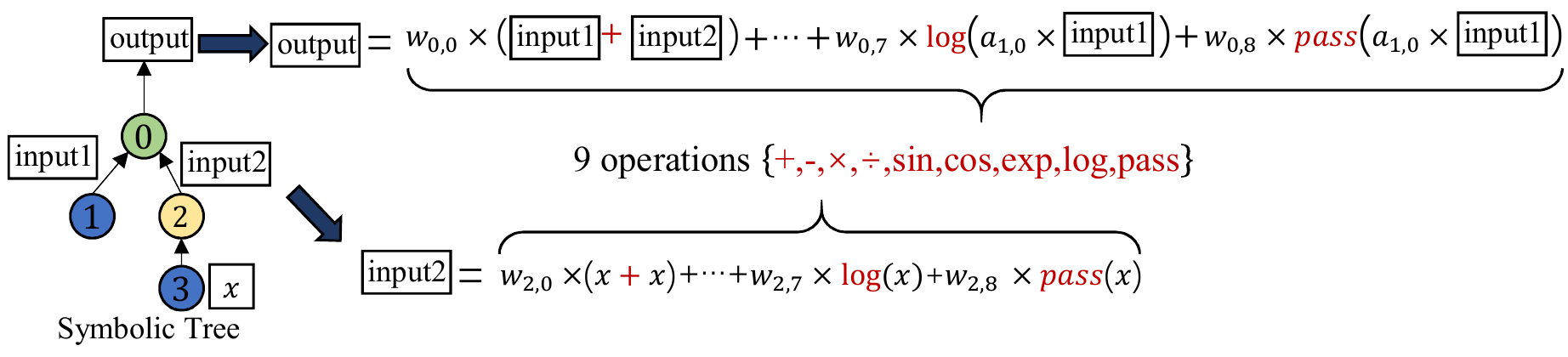}
	\caption{An example to explain how the fixed node in the symbolic tree is relaxed to a continuous representation.} 
	\label{mixing_node}
\end{figure*}

Given the node matrix $N$ and the adjacency matrix $A$, we can show how the fixed node in the original symbolic tree is relaxed to a continuous representation. Overall, there are two kinds of nodes (2-arity and 1-arity) in the symbolic tree, corresponding to two different situations:

1$)$ For a 2-arity node: with the proposed $N$ and $A$, a 2-arity node is transformed to a mixing node of all the operations. Specifically, the mixing node takes the outputs of two child nodes as inputs. For binary operations, such as $+$, $-$, the node calculates the results of each binary operation with the two inputs and then weights the results using the corresponding weights in $N$. For unary operations, such as $sin$, $exp$, they only need one input. Therefore, the child node with the greater adjacency weight in $A$ would be chosen and the output of it is taken as the input. The results of each unary operation are calculated with the input and then weighted with the corresponding weights in $N$. Finally, the sum of all the results is returned as the output of this node. 

2$)$ For a 1-arity node: a 1-arity node is also transformed to a mixing node of all the operations. Specifically, a 1-arity node takes the only child node as the original input. For unary operations, the node calculates the results of each unary operation with the original input and returns the weighted sum. For binary operations, a leaf node would be added as the second input for the node, and the results of each binary operation are calculated using the two inputs. Through the above approach, nodes in the tree are transformed into mixing nodes, allowing continuous representation of the symbolic tree to be obtained. 

As previously discussed, bloating is a common problem in GP-based SR, which also occurs in the proposed method. In order to alleviate this issue, a new unary operation called $pass$ is introduced, to indicate that the input of a node is returned directly as the result. The effect of $pass$ is taken in conjunction with the \textit{Diversification} part designed in the proposed method, and the details are justified in Subsection~\ref{diversification}.

For a better of understanding the above design, we provide an example in Fig.~\ref{mixing_node} to explain how the nodes 0 and 2 are transformed into mixing nodes. Based on the design of $pass$ and also the 8 operations defined above, there are a total of 9 operations mixed in the nodes. In the following, we take the operations $+$ and $pass$ as the representatives of binary operations and unary operations, respectively, for the explanations. Specifically, the node 0 is a 2-arity node that takes the outputs of two child nodes as inputs, as marked by input1 and input2. For the binary operation $+$, the corresponding weight is $w_{0,0}$ in $N$. Therefore, we calculate the result of $+$ with input1 and input2, and weigh the result with $w_{0,0}$. For the unary operation $pass$, the corresponding weight is $w_{0,8}$ and it only needs one input. Assuming the adjacency weight of input1 $a_{1,0}$ is greater than that of input2, we calculate the result of $pass$ with the product of $a_{1,0}$ and input1, and weigh the result with $w_{0,8}$. Furthermore, the node 2 is a 1-arity node, and it takes the output of the leaf node 3 marked by $x$ as input. For the binary operation $+$, the corresponding weight is $w_{2,0}$ and it needs two inputs. Therefore, we add a leaf node as the second input, and calculate the result of $+$ with the two inputs and weigh the result with $w_{2,0}$. For the unary operation $pass$, it calculates the result of $pass$ with the input $x$ and weights it with $w_{2,8}$, which is the weight for $pass$ in $N$. Through the above process, the node 0 and node 2 are transformed into the mixing nodes.

Lastly, in order to take advantage of the previously learned information, $N$ and $A$ are initialized according to the original symbolic tree rather than through a random process. Taking the symbolic tree in Fig.~\ref{gradient_descent} as an example, the primitive at the root node of the symbolic tree is $+$, and so the weight of $+$ is initialized to 1, with others set to 0 in the node vector of the root node when constructing $N$.

The symbolic tree, the node matrix $N$, and the adjacency matrix $A$ together form the model called the differentiable symbolic tree (DST). Through the proposed DST, the structure of a symbolic tree is represented by the corresponding real-valued matrices $N$ and $A$, which can then be optimized using gradient-based optimization.

\subsubsection{Gradient-based Optimizer}
After mapping the symbolic tree into a continuous space, a gradient-based approach can be easily used to optimize the structure. We approximate the search of the symbolic tree structure as a differentiable problem shown in Equation~(\ref{approximate}):
\begin{equation}
	\left\{
	\begin{aligned}
		F^{*} &= arg\ min_{F\in Q}L(F(X), y) \\
		&\xrightarrow[]{F\ is \ determined \ by \ N,A} \\
		N^{*}, A^{*} &= arg\ min_{N, A} L(F_{N, A}(X), y) \label{approximate}
	\end{aligned}\right.
\end{equation}
where $F$ represents the mathematical expressions in $Q$, and $Q$ is the whole search space defined by the terminal and function sets. $X\in\mathbb{R}^{|t|}$ and $y\in\mathbb{R}$ represents the data points in the training dataset. $L$ is the predefined loss function, which measures the closeness of $F(X)$ to the target $y$. The first line in Equation~(\ref{approximate}) is for the discrete space, where a genetic beam search is used to optimize $F$ towards the optimal expression $F^{*}$ based on the training dataset. After mapping the mathematical expression to the DST, $F$ is determined by two data structures: $N$ and $A$ (the second line in Equation~(\ref{approximate})). Thus, the problem transforms to one of making the output of the DST model close to $y$ by adjusting $N$ and $A$. Hence, the DST can be optimized in a continuous domain using a gradient-based optimization approach. This optimization approach consists of three parts: forward propagation, loss function, and gradient descent optimization.

\textbf{Forward propagation:} As in a symbolic tree, the DST takes the input variables in the dataset as input and returns the computed result of the root node as output. The forward propagation process of DST consists of two parts: the mixing computation within nodes and the propagation of information between nodes.

For the mixing computation within nodes, let $O$ be a set of candidate operations of the node, $O_{j}(\cdot)$ be the j-th operation to be applied to the input of the node $x^{(i)}$, $w_{i,j}$ be the weight of the i-th row and j-th column in $N$, and $L$ be the length of $O$. In forward propagation process, the output of the i-th node $\bar{O}(x^{(i)})$ is a weighted sum over all possible operations, as shown in Equation~(\ref{softmax}).
\begin{equation}
	\bar{O}(x^{(i)}) = \sum_{j=0}^{L-1} w_{i,j}O_{j}(x^{(i)}) \label{softmax}
\end{equation}
Through this equation, the output of a node in the DST is computed and then provided to the next node as an input.

\begin{algorithm}
	\DontPrintSemicolon
	
	\KwInput{The input data $X$, the node matrix $N$, the adjacency matrix $A$, the origin tree $T$, the operation set $Op$}
	\KwOutput{The output of DST $ y_{pred} $} 
	$n_{node} \leftarrow$ Compute the number of nodes in $T$\; \label{a2_l1}
	$r \leftarrow$ Store the calculation results of all the nodes\; \label{a2_l2}
	
	\For{id in reverse($n_{node}$)}{ \label{a2_l3}
		$n \leftarrow$ Get the node in $T$ according to $id$\; \label{a2_l4}
		$w \leftarrow$ Get the node weight in $N$ according to $id$\; \label{a2_l5}
		\If{$n$ is leaf\_node}{ \label{a2_l6}
			$r[id] \leftarrow$ Compute the results of $n$ with $w$, $Op$ and $X$ according to Equation~(\ref{softmax})\; \label{a2_l7}
		} 
		\Else{ \label{a2_l8}
			$a \leftarrow$ Get the adjacency weight in $A$ according to $id$\; \label{a2_l9}
			$id_{child} \leftarrow$ Obtain the childs of the node according to $a$\; \label{a2_l10}
			$X^{(id)} \leftarrow$ Compute the inputs according to $a$ and $r[id_{child}]$\; \label{a2_l11}
			$r[id] \leftarrow$ Compute the results of $n$ with $w$, $Op$, $X^{(id)}$ according to Equation~(\ref{softmax})\; \label{a2_l12}
		}
	}
	$y_{pred} \leftarrow r[0]$ \; \label{a2_l13}
	\textbf{Return} $y_{pred}$\; \label{a2_l14}
	
	\caption{The Forward Propagation Process of DST}
	\label{algorithm2}
\end{algorithm}

Our proposed approach uses the adjacency matrix to define the propagation of information between nodes, as described formally in Algorithm~\ref{algorithm2}. A bottom-up execution process is used, with an array ($r$) storing the computation results of previous nodes (lines~\ref{a2_l1}\&\ref{a2_l2}). Starting from the leaf nodes, for each node in the DST, we first obtain the node's weight from $N$ (lines~\ref{a2_l3}-\ref{a2_l5}), and then compute its output based on the type of the node. For a leaf node, its inputs are directly read from the training dataset, with its output computed based on Equation~\ref{softmax} (lines~\ref{a2_l6}\&\ref{a2_l7}). For a 2-arity node, the two child nodes, as well as their adjacency weights, are found from $A$ (lines~\ref{a2_l9}\&\ref{a2_l10}). The weighted outputs of the two child nodes are fed as inputs to the current node for its calculation (lines~\ref{a2_l11}\&\ref{a2_l12}). The process for a 1-arity node is similar, but only one adjacency weight and the corresponding node are selected. After finishing the computation for all the nodes in the DST, the result of the root node is returned as the output of the DST (lines~\ref{a2_l13}\&\ref{a2_l14}).

\textbf{Loss function:} In GP-based SR, a standard fitness measure is the normalized root-mean-square error (NRMSE)~\cite{ONeill.2009} (where the standard RMSE is normalized by the standard deviation of the target values $y$). That is, given a dataset consisting of $n$ groups of $(X, y)$ pairs, NRMSE is computed as Equation~(\ref{nrmse}):
\begin{equation}
	NRMSE =\frac{1}{\sigma}\sqrt{\frac{1}{n}\sum^{n}_{i=1}(y_{i}-\hat{y_{i}})^{2}} \label{nrmse}
\end{equation}
where $\hat{y} = f(X)$ are the predicted values computed using the candidate expression $f$, and $\sigma$ is the standard deviation of the target values $y$. Normalization by $\sigma$ makes the measure consistent across different datasets which have different domains. We use this standard measure as our loss function.

Given that we are performing SR, our DGP method needs to reacquire the discrete structure from the continuous encoding. However, there are possible issues when discretizing the continuous encodings. The weight vector of certain nodes may not clearly specify a single operation. For example, if the weight vector of a node is $[0.312, 0.135, ..., 0.315]$, it is hard to justify that an operation weighted by $0.315$ is significantly better than the other weighted by $0.312$. To address this, an extra loss is explicitly incorporated to push values in the node matrix towards 0 or 1 (as per~\cite{Chu.2020}), formally as Equation~(\ref{loss0-1}):
\begin{equation}
	Loss_{0-1} = -\frac{1}{L}\sum_{j=1}^{L}(w_{j}-0.5)^{2} \label{loss0-1}
\end{equation}
where $L$ is the number of operations mixed in a node, and $w$ is the softmax value of the weight of different operations. In order to control the strength of $Loss_{0-1}$, the loss is weighted by a coefficient $\lambda_{0-1}$. Thus, the final loss function is formulated as Equation~(\ref{loss}):
\begin{equation}
	Loss = NRMSE + \lambda_{0-1}Loss_{0-1} \label{loss}
\end{equation}

\textbf{Gradient descent optimization:} The gradient descent process for optimizing $N$ and $A$ is described in Algorithm~\ref{algorithm3}. In each epoch, a batch of data is first obtained from the training dataset in the form $(X, y)$ (line~\ref{l4}). The input data $X$ is fed into the DST model, and the forward propagation results of DST are obtained as the predicted values $y_{pred}$ (line~\ref{l5}). The loss between $y$ and $y_{pred}$ is then computed according to Equation (\ref{loss}) (line~\ref{l6}). After that, the gradients of the learnable parameters in the $N$ and $A$ are computed by back-propagation based on the loss (line~\ref{l7}). It should be noted that since the operations mixed in the nodes are involved in the computation of the forward propagation results, the gradients of the intermediate mixing nodes are also computed in the process of back-propagation. These gradients are then applied to the $N$ and $A$ to fine-tune the learnable parameters (lines~\ref{l8}\&\ref{l9}). The above process is performed over multiple epochs to reduce the loss iteratively. Finally, the optimized matrices from the DST are returned(lines~\ref{l10}\&\ref{l11}).
\begin{algorithm}[!htp]
	\DontPrintSemicolon
	
	\KwInput{The DST model $M$, the dataset $D$, the maximal epochs $E$, the loss function $L$, the batch size $n$, the learning rate $\alpha$ and $\beta$}
	\KwOutput{The optimized parameter matrices $N_{opt}$ and $A_{opt}$}
	\For{epoch = 1, ..., $E$}{ \label{l3}
		$(X_{i}, y_{i})_{i=1}^{n}\leftarrow$Get a batch of data (input-output pairs) from $D$\; \label{l4}
		$y_{pred} \leftarrow M(X)$\; \label{l5}
		$loss \leftarrow L(y, y_{pred})$\; \label{l6}
		$grad_{N}$, $grad_{A}\leftarrow$Compute the gradients of $N$ and $A$ by back-propagation of $loss$\; \label{l7}
		N $\leftarrow$ N + $\alpha grad_{N}$\; \label{l8}
		A $\leftarrow$ A + $\beta grad_{A}$\; \label{l9}
	}
	$N_{opt}$, $A_{opt} \leftarrow$ Get the optimized parameter matrices from the trained model $M$\; \label{l10}
	\textbf{Return} $N_{opt}$, $A_{opt}$ \label{l11}
	\caption{Gradient Descent Optimization for the DST}
	\label{algorithm3}
\end{algorithm}

The above three parts (forward propagation, the loss function and gradient descent optimization) together form the gradient-based optimizer. In each iteration, the expressions from the previous iteration are first transformed into symbolic trees and then relaxed to generate DSTs. Each DST is fed into the optimizer and is iteratively trained on the dataset until convergence. Then, the optimized DSTs go through a \textit{Diversification} process, as introduced in the next subsection.

\subsection{Diversification} \label{diversification}
The Diversification phase is proposed to drive the search into diverse subspaces. The goal is achieved by first sampling symbolic trees from the optimized DST, and then applying a diversifier to generate new symbolic trees as the starting point of the next \textit{Optimization} phase. In the following, these two steps are introduced in detail.

First, new symbolic trees are sampled from the optimized DST. The sampling step also uses a bottom-up approach. Starting from the leaf node, for each node in the DST, one of the primitives is selected with probability based on the optimized $N$. We develop three operations: SHIRINK, REPLACE, and EXPAND for the sampling step. Depending on the change of the primitive in the node, one of the three operations is chosen to perform on the node. The SHRINK operation is carried out if the arity of the node decreases; the REPLACE operation is carried out if the arity is not changed; and the EXPAND operation is carried out if the arity increases. These operations are as follows.

\begin{itemize}
	\item SHRINK: deleting the node. If the node is unary, the child node will now directly connect to the parent node. Otherwise, the child node with the greater adjacency weight will be chosen to connect to the parent node. 
	\item REPLACE: replacing the node with a new function or terminal node, but not changing the number of arity in the node. 
	\item EXPAND: changing the type of the node, e.g., a leaf node to a 1-arity node or a 1-arity node to a 2-arity node, and connecting new children nodes to it. 
\end{itemize}

\begin{figure}[!htp]
	\begin{center}
		\subfloat[The SHRINK operation]{\includegraphics[width=0.5\columnwidth]{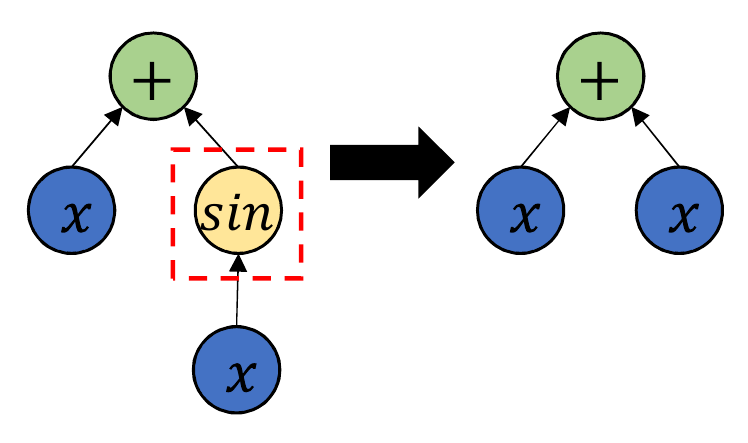}%
			\label{fig_SHRINK_operation}}
		\hfil
		\subfloat[The REPLACE operation]{\includegraphics[width=0.45\columnwidth]{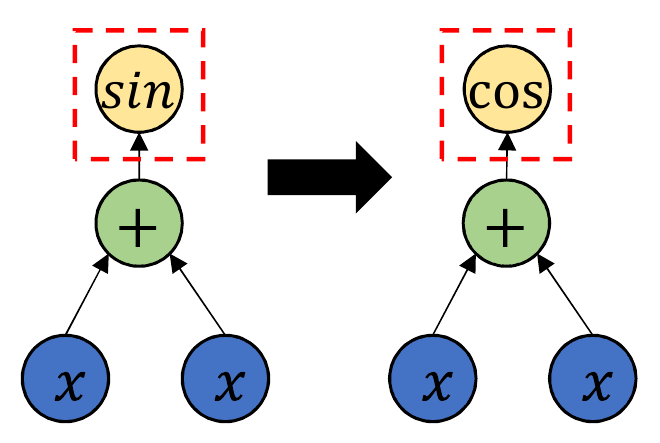}%
			\label{fig_REPLACE_operation}}
		\hfil
		\subfloat[The EXPAND operation]{\includegraphics[width=0.5\columnwidth]{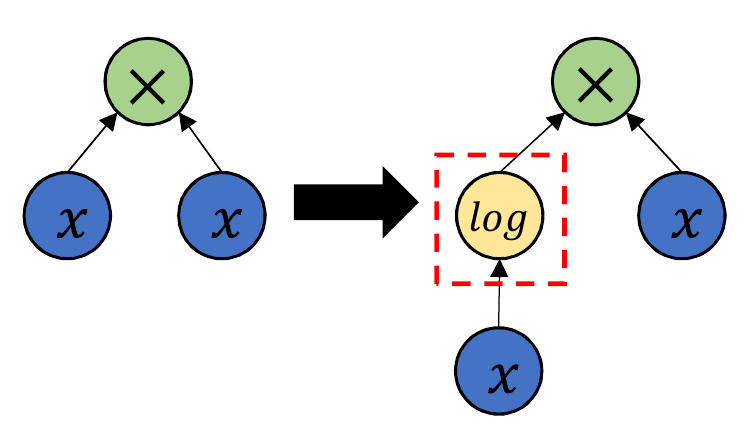}%
			\label{fig_EXPAND_operation}}

		\caption{Examples of the SHRINK, REPLACE, and EXPAND operations. The node(s) targeted by the corresponding operation has been surrounded with red dash rectangle for the purpose of illustration.}
		\label{node_operation}
	\end{center}
\end{figure}

Please note that the REPLACE operation is similar to the EXPAND operation, where each node can be changed by another node, no matter whether the node is a terminal or a function. They differ in that the REPLACE operation requires the number of arity unchanged, while the EXPAND operation does not. Fig.~\ref{node_operation} illustrates the examples of these three operations. In particular,  in Fig.~\ref{node_operation} (a), the $sin$ node is shrunk, and its child node is directly connected to its parent node; in Fig.~\ref{node_operation} (b), the function $sin$ in the root node is replaced by a new function $cos$; in Fig.~\ref{node_operation} (c), the left child of the root node is expanded. The primitive in it is changed from $x$ to $log$, and a leaf node is added to connect with it. 

It has been highlighted that the design of $pass$ can alleviate the bloating issue faced by existing GP methods. In the proposed method,  if $pass$ accounts for the maximum weight in a node, the node would be removed to shrink the symbolic tree, reducing the presence of bloating. This is because the introduction of pass gives a way to reduce the depth of the symbolic tree. Through the pass operation, the redundant nodes in the symbolic tree can be removed and the bloating can be effectively controlled.

By performing the above three operations, we are able to sample well-behaved symbolic trees from the subspace of the symbolic tree. However, the sampled symbolic trees are only the local optimal solutions in the current subspace, and a mechanism is needed to guide the search towards potentially better subspaces. Inspired by recent work~\cite{Mundhenk.2021}, we designed a diversifier to tackle the problem. The diversifier applies common genetic operators including crossover and mutation to the sampled symbolic trees to generate more diverse structures. The use of evolutionary operations generates individuals that may fall well outside the local region of the search space and can effectively avoid the search process falling into a local optimum.

Finally, the new symbolic trees are evaluated on the training dataset, and their accuracy is computed. If their accuracy reaches a pre-defined threshold, we consider that we have found the correct mathematical expression, and the search is ended. Otherwise, the next iteration of optimization and diversification restarts based on the current symbolic trees. The above process continues until the maximum number of generations is reached, and then the best expression found is returned as the final result. 

\section{Experiment Design}\label{experiment}
In this section, the experiment settings of the proposed method as well as other baseline methods are given. The benchmark datasets are detailed in Subsection~\ref{Datasets}. The GP-based and NN-based SR recently-proposed baselines are introduced in Subsection~\ref{Baseline Methods}. The evaluation metrics and parameter settings are shown in Subsection~\ref{Parameter Settings}.

\subsection{Benchmark Datasets}\label{Datasets}
To verify if the proposed DGP method can effectively and efficiently solve high-dimensional SR problems, we chose eight real-world datasets (listed in Table \ref{realwork_benchmark}) that are commonly used in the SR community~\cite{Chen.2022,Franca.2021,LaCava.2021}. The first seven datasets are taken from the Penn Machine Learning Benchmarks (PMLB)~\cite{Olson.2017}, which is a large collection of curated benchmark datasets for evaluating and comparing supervised machine learning algorithms. These are the regression benchmarks with the highest number of features in PMLB, ranging from tens to hundreds of features, which is challenging for both GP-based and NN-based SR methods. In order to investigate the effect of the proposed DGP on even higher-dimensional problems, we added the last benchmark (DLBCL) to the experiments. The DLBCL dataset is a diffuse large B-cell lymphomas dataset with over $7,000$ features, which is hundreds of times more than commonly-used regression datasets. Each dataset is divided into training and test sets at a ratio of 75\% to 25\%.

\begin{table}[!htbp]
	\centering
	\caption{Real-world Benchmarks}
	\renewcommand{\arraystretch}{1.2}
	\begin{tabularx}{0.5\textwidth}{X|l|l|l|l}
		\hline
		Name                  &\#Features &\#Total&\#Training &\#Test  \\ \hline \hline
		Satellite\_image              & 36  & 6,435 & 4,826 & 1,609\\
		Fri\_c4\_50           & 50  & 1,000 & 750 & 250\\ 
		Fri\_c0\_50           & 50 & 250 &187 & 63\\
		Fri\_c1\_50          & 50  & 500   & 375 & 125 \\
		Fri\_c4\_100          & 100 & 1,000 & 750 & 250 \\
		GeoOriMusic           & 117 & 1,059 & 794 & 265\\ 
		Tecator               & 124 & 240 & 180 & 60 \\	
		DLBCL                            & 7,400 & 240  & 180   & 60\\ \hline
	\end{tabularx}
	\label{realwork_benchmark}
\end{table}

Another common practice in the SR community is to verify if the exact regression equations can be found by SR methods. To this end, we evaluate DGP and its competitors on six synthetic SR benchmarks, which are commonly used in previous research~\cite{Petersen.2021,Biggio.2021,Mundhenk.2021}. This is because the ground truth expressions of the synthetic benchmarks are pre-defined. The details of the six benchmarks are listed in Table~\ref{synthetic_benchmark}. The data points for training and testing are uniformly sampled 20 times, as in previous papers~\cite{Petersen.2021,Mundhenk.2021}. 

\begin{table}[!htbp]
	\centering
	\caption{Synthetic Benchmarks}
	\renewcommand{\arraystretch}{1.2}
	\begin{tabularx}{0.5\textwidth}{X|l|l}
		\hline
		Truth Expression      &Denoted &Input Range  \\ \hline \hline
		$sin(x^{2}) cos(x) - 1$              & S1  & $(-1, 1)$ \\
		$log(x + 1) + log(x^{2} + 1)$           & S2  & $(0, 2)$\\ 
		$x^{3} + x^{2} + x + sin(x) + sin(x^{2})$           & S3 & $(-1, 1)$ \\
		$sin(x) + sin(y^{2})$           & S4  & $(0, 1)$  \\
		$\frac{x^{4}}{x+y}$         & S5 & $(-1, 1)$ \\
		$4 \times sin(x) cos(y)$           & S6 & $(0, 1)$\\  \hline
	\end{tabularx}
	\label{synthetic_benchmark}
\end{table}

\subsection{Peer Competitors}\label{Baseline Methods}
We chose recently proposed SR methods with good performance as our baseline methods. Both GP-based and NN-based methods are included:

\begin{itemize}
	\item Genetic Programming (GP): the standard GP-based symbolic regression implemented by the evolutionary computation framework DEAP~\cite{Fortin.2012}.
	\item Rademacher Complexity for Enhancing the Generalization of Genetic Programming for Symbolic Regression~\cite{Chen.2022} (MGPRC): A recently proposed multiobjective GP-based symbolic regression method equipped with a novel complexity measure based on the Rademacher complexity. The complexity of an evolved model is measured by the maximum correlation between the model and the Rademacher variables on the selected training instances. By minimizing the training error and the Rademacher complexity of the models as the two objectives, MGPRC has shown to be superior to the standard GP on generalization performance. 
	\item Multi-Objective Genetic Programming Using NSGA-II for Symbolic Regression~\cite{Virgolin.2020} (NSGP): A new model of interpretability is built and incorporated into the GP algorithm. Through the approach, NSGP can achieve a better  accuracy-interpretability trade-off. In addition, the penalization of duplicates~\cite{Virgolin.2021} is added to the method to preserve diversity better and find more accurate models.
	\item Deep Symbolic Regression~\cite{Petersen.2021} (DSR): a recently proposed deep learning method for symbolic regression based on RNN generating symbolic trees and a modified reinforcement learning approach named risk-seeking policy gradient is utilized to optimize the RNN.
	\item Symbolic Regression via Neural-Guided Genetic Programming Population Seeding~\cite{Mundhenk.2021} (PSSR): a hybrid RNN and GP approach to symbolic regression. A neural-guided component is used to seed the starting population of a random restart genetic programming component, gradually learning better starting populations.
\end{itemize}

\subsection{Evaluation Metrics and Parameter Settings}\label{Parameter Settings}
In order to reasonably compare the performance of different methods on real-world and synthetic benchmarks, different evaluation metrics are used. For real-world benchmarks, we follow the commonly evaluation metric~\cite{LaCava.2021} and compare the coefficient of determination $R^{2}$ between $y$ and $\hat{y}$, excluding NaNs and $\pm\infty$. The $R^{2}$ result is define as Equation (\ref{r2}):

\begin{equation}
	R^{2} = 1- \frac{\sum _{i}^{N}(y_{i}-\hat{y}_{i})^{2}}{\sum_{i}^{N}(y_{i}-\bar{y}_{i})^{2}} \label{r2}
\end{equation}
The introduction of the metric removes the interference of very few outliers in the evaluation results and reflects more effectively the merits of the searched symbolic expressions. For the synthetic benchmarks with exact expressions, we further compared the recovery rates~\cite{Petersen.2021} between the different methods. Recovery rate is defined as the number of runs that successfully recovered the expression from the data as a percentage of the total number of runs. The metric can effectively eliminate the uncertainty of performance evaluation brought by randomness.

In the experiments, each method is trained on each dataset across 30 repeated trials with a different random state that controls the seed of the method. The parameter settings of the experiments are in Table~\ref{experiment setting}. Different parameters are used in the experiments for NN-based and GP-based SR methods. Given that it is difficult to establish an exact correspondence between GP-based and NN-based methods, e.g., how many training epochs in NN-based methods correspond to one evolution generation in GP, we follow the settings in~\cite{LaCava.2021} and uniformly set the termination criteria as the number of evaluations performed. We set the maximum number of evaluations to 100k for real-world benchmarks and 500k for synthetic benchmarks. 
\begin{table}[!ht]
	
	\caption{Hyperparameter settings for the experiments.}
	\renewcommand{\arraystretch}{1.2} 
	\begin{tabular}{l|l}
		\hline
		\textbf{Hyperparameter}  & \textbf{Value}\\ \hline
		\hline
		\textbf{Common parameters}&   \\ \hline
		Function set  & $+$, $-$, $\times$, $\div$, $sin$, $cos$, $exp$, $log$ \\ \hline
		Termination criteria  & 100k evaluations$^{\dagger}$\\\hline
		Number of repeated trials & 30$^{\dagger}$ \\ \hline
		\hline
		\textbf{Parameters of the proposed DGP}&  \\ \hline
		Optimizer  &  Adam  \\ \hline
		Learning rate  &0.005  \\ \hline
		Loss function &NRMSE + 0.1$Loss_{0-1}$ \\ \hline
		Population size  & 500$^{\dagger}$  \\ \hline
		Epochs per iteration & 1,000 \\ \hline
		Crossover \& mutation rates & 0.5\&0.5 \\ \hline
		Generations per iteration  & 20 \\ \hline
		\hline
		\textbf{Parameters of NN-based methods} &    \\ \hline
		Optimizer  &  Adam  \\ \hline
		RNN cell type   & LSTM \\ \hline
		RNN cell layers   &  1 \\ \hline
		RNN cell size   & 32 \\ \hline
		Training method  &RSPG \\ \hline
		Learning rate  & 0.0025  \\ \hline
		Entropy weight  & 0.005  \\ \hline
		Minimum expression length  & 4  \\ \hline
		Maximum expression length &30  \\ \hline
		Reward/fitness function  & NRMSE  \\ \hline
		\hline
		\textbf{Parameters of GP-based methods}  &  \\ \hline
		Population size  & 500$^{\dagger}$  \\ \hline
		Initialization method  & Ramped half-and-half  \\ \hline
		Crossover operator  & One point  \\ \hline
		Crossover probability  & 0.5  \\ \hline
		Mutation operator  & Uniform \\ \hline
		Mutation probability & 0.5  \\ \hline
		Selection operator  & Tournament  \\ \hline
		Tournament size  & 3  \\ \hline
		Mutate tree maximum & 2  \\ \hline
		Mutate tree minimum & 0  \\ \hline
	\end{tabular}
	$^{\dagger}$ The termination criteria, the number of repeated trials, and the population size for synthetic experiments are 500k, 10, and 1,000, respectively.
	\label{experiment setting}
\end{table}

\section{Results and Discussion}\label{result}
The experiments in this section are divided into two groups: real-world and synthetic benchmarks. On the real-world benchmarks, we measure training and generalization performance~\cite{Chen.2022,Franca.2021,Chen.2019}. On the synthetic benchmarks, following previous research~\cite{Petersen.2021,Mundhenk.2021}, we perform experiments to determine the recovery rate and the effect of noise. As discussed in Section~\ref{sec:introduction}, existing GP methods suffer from the problem of bloat on SR problems, resulting in very large and un-interpretable expressions. To compare the ability of the proposed method to address this issue, we also perform program size analysis of the final expressions across different methods in both real-world and synthetic experiments.

\subsection{Results on the Real-world Benchmarks} \label{realworld subsection}
\begin{figure*}[htp]
	\begin{center}
		\subfloat[Satellite\_image]{\includegraphics[width=0.5\columnwidth]{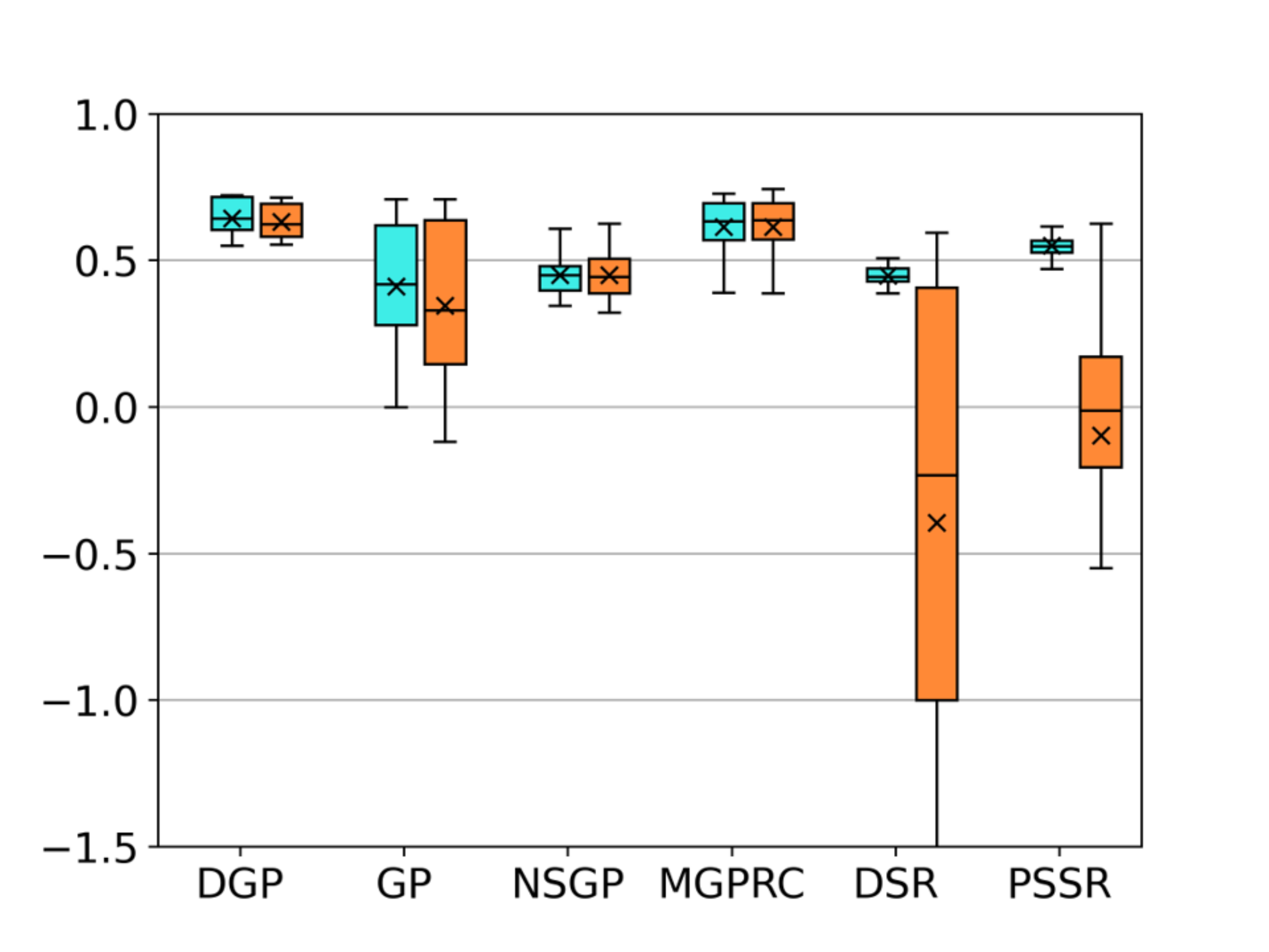}\label{fig_Satellite_image_result}}
		\hfil
		\subfloat[Fri\_c4\_50]{\includegraphics[width=0.5\columnwidth]{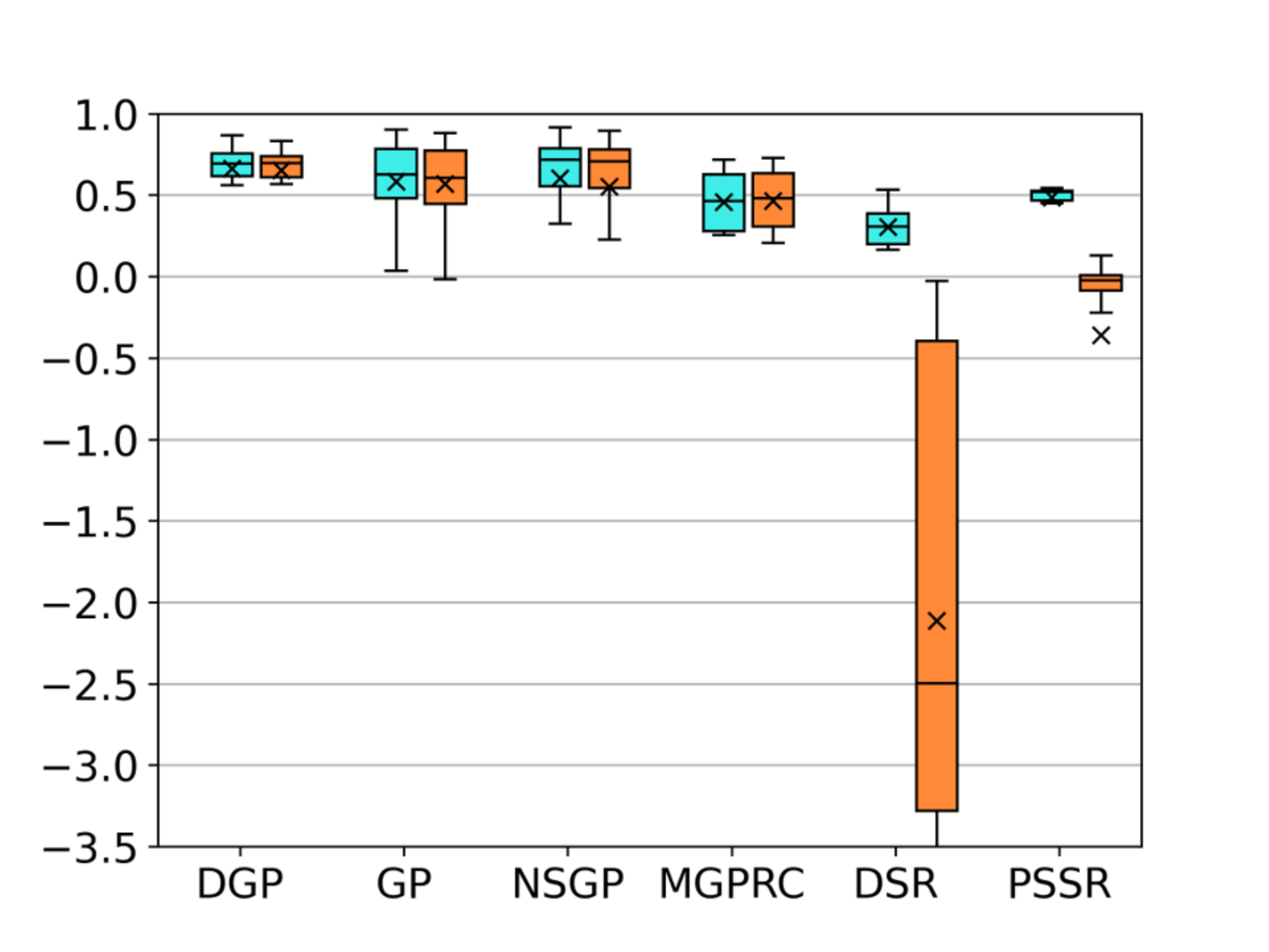}\label{fig_Fri_c4_50_result}}
		\hfil
		\subfloat[Fri\_c0\_50]{\includegraphics[width=0.5\columnwidth]{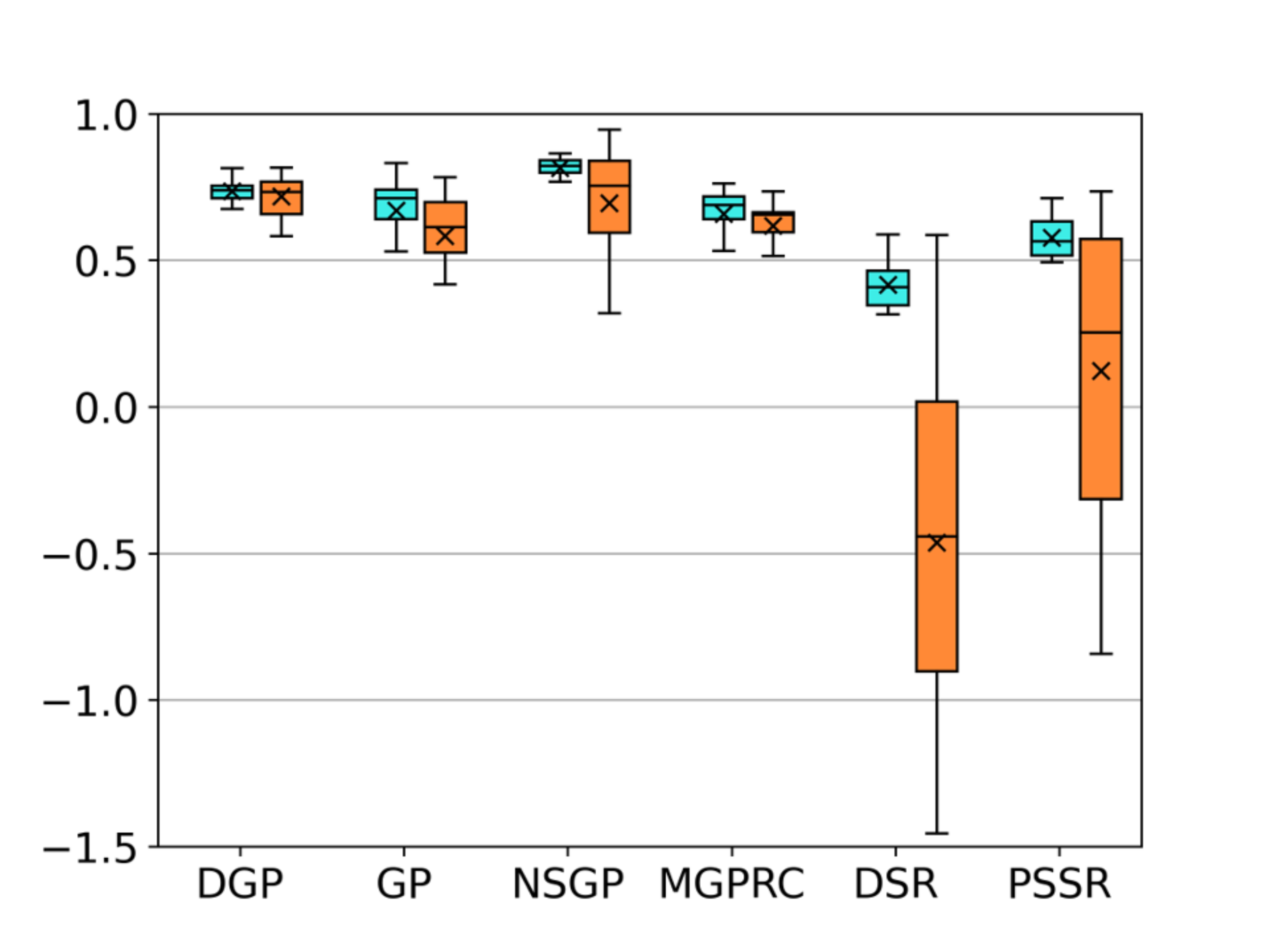}\label{fig_Fri_c0_50_result}}
		\hfil
		\subfloat[Fri\_c1\_50]{\includegraphics[width=0.5\columnwidth]{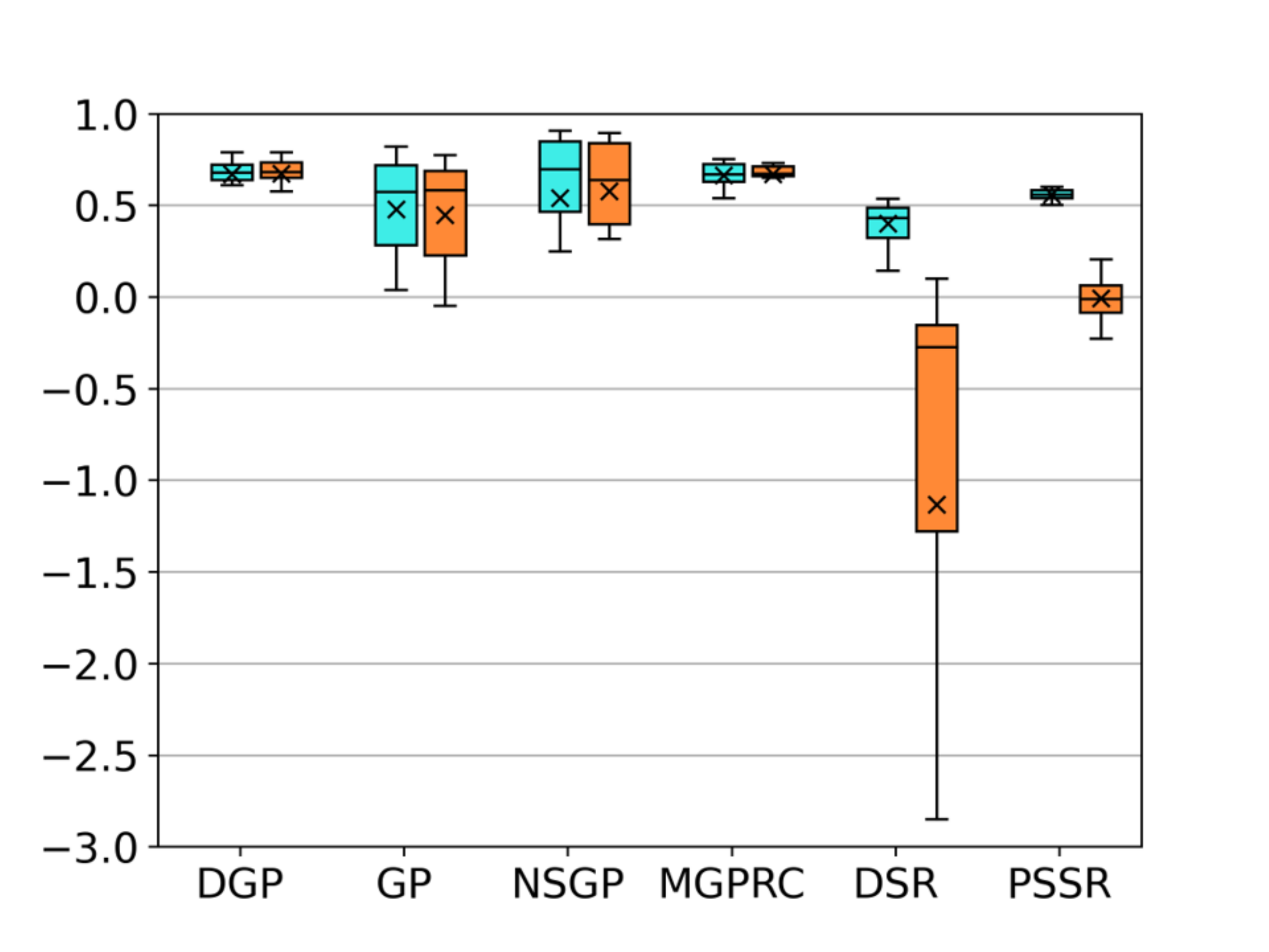}\label{fig_Fri_c1_50_result}}
		\hfil
		\subfloat[Fri\_c4\_100]{\includegraphics[width=0.5\columnwidth]{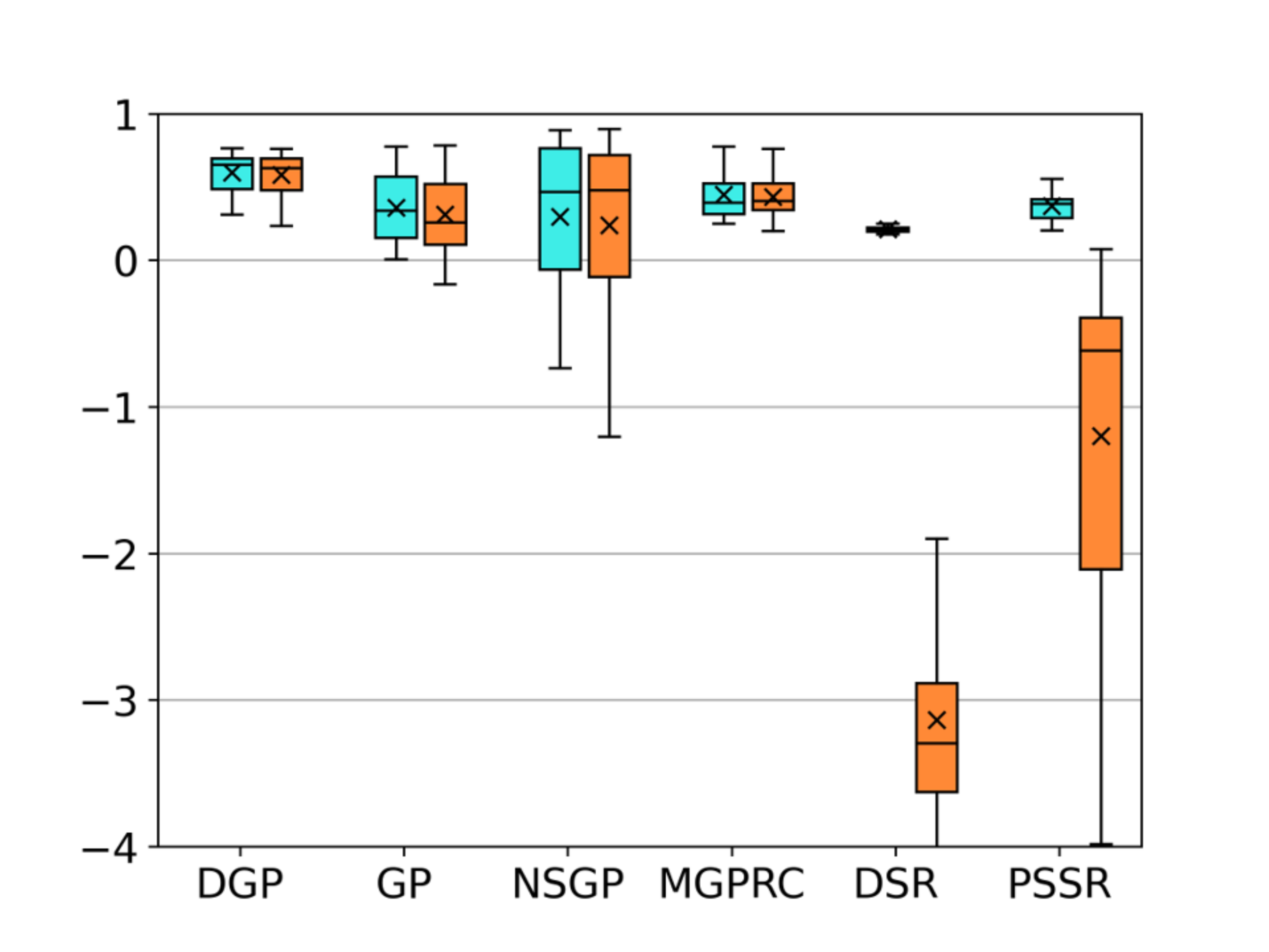}\label{fig_Fri_c4_100_result}}
		\hfil
		\subfloat[GeoOriMusic]{\includegraphics[width=0.5\columnwidth]{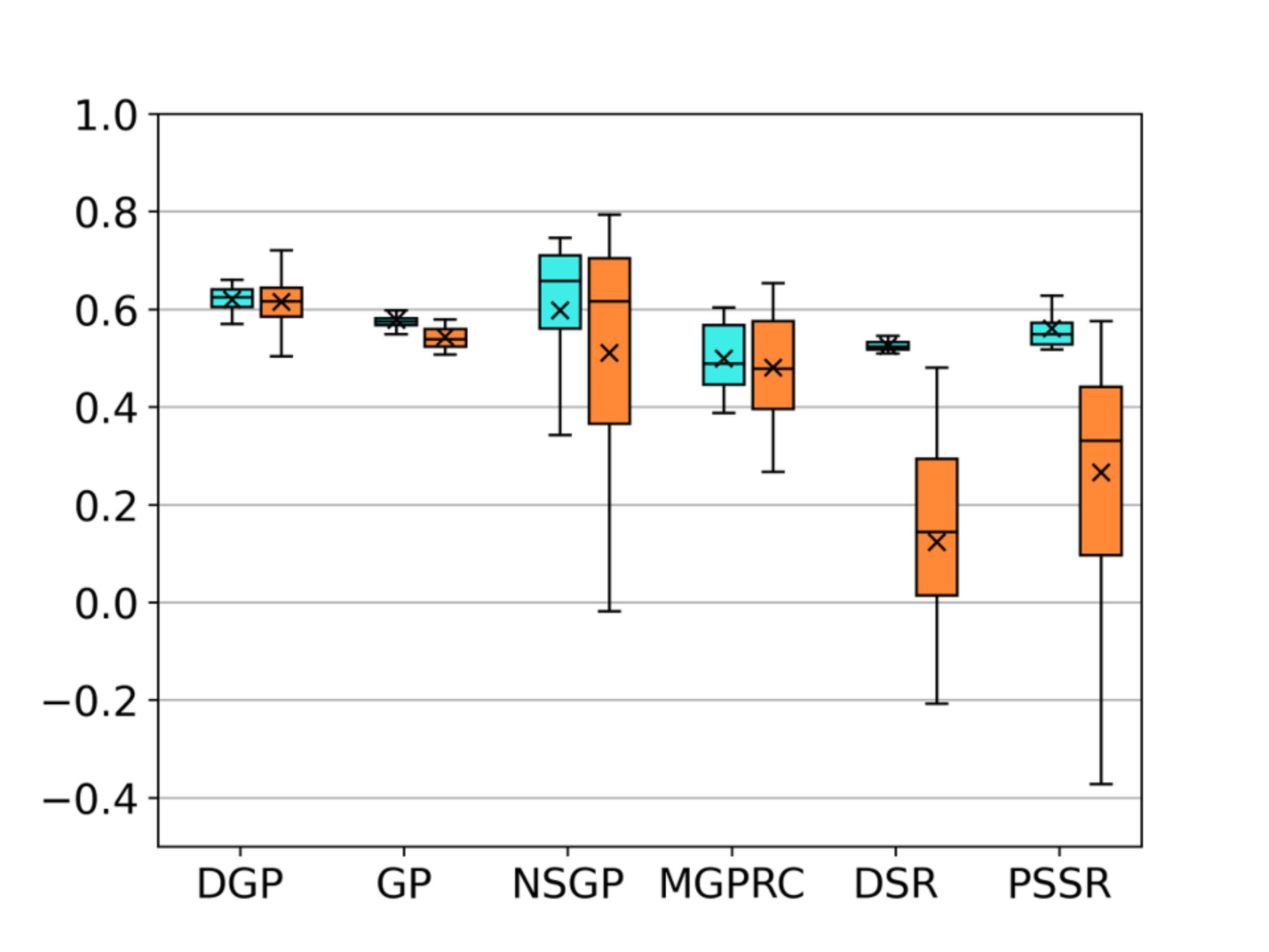}\label{fig_GeoOriMusic_result}}
		\hfil
		\subfloat[Tecator]{\includegraphics[width=0.5\columnwidth]{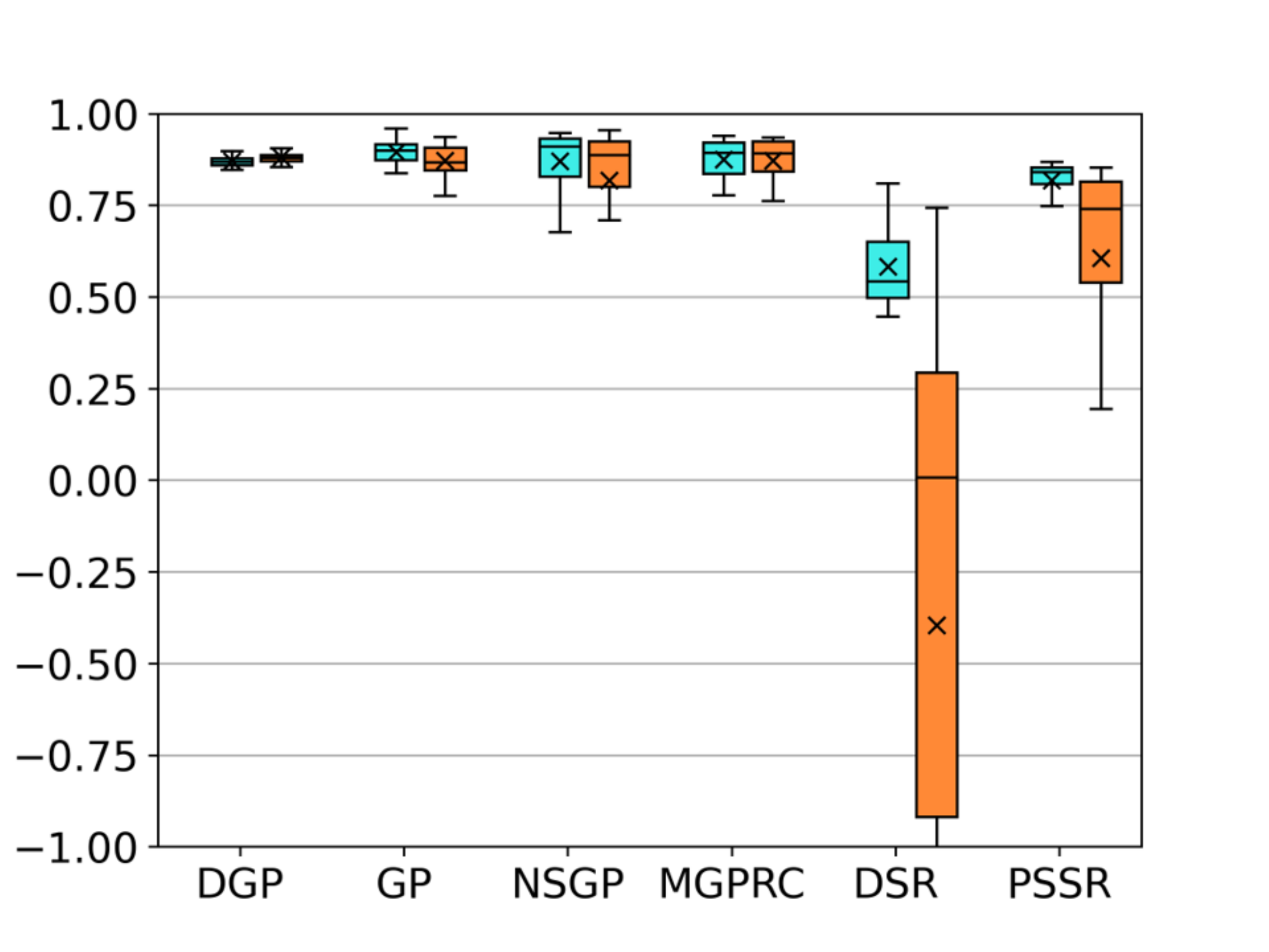}\label{fig_Tecator_result}}
		\hfil
		\subfloat[DLBCL]{\includegraphics[width=0.5\columnwidth]{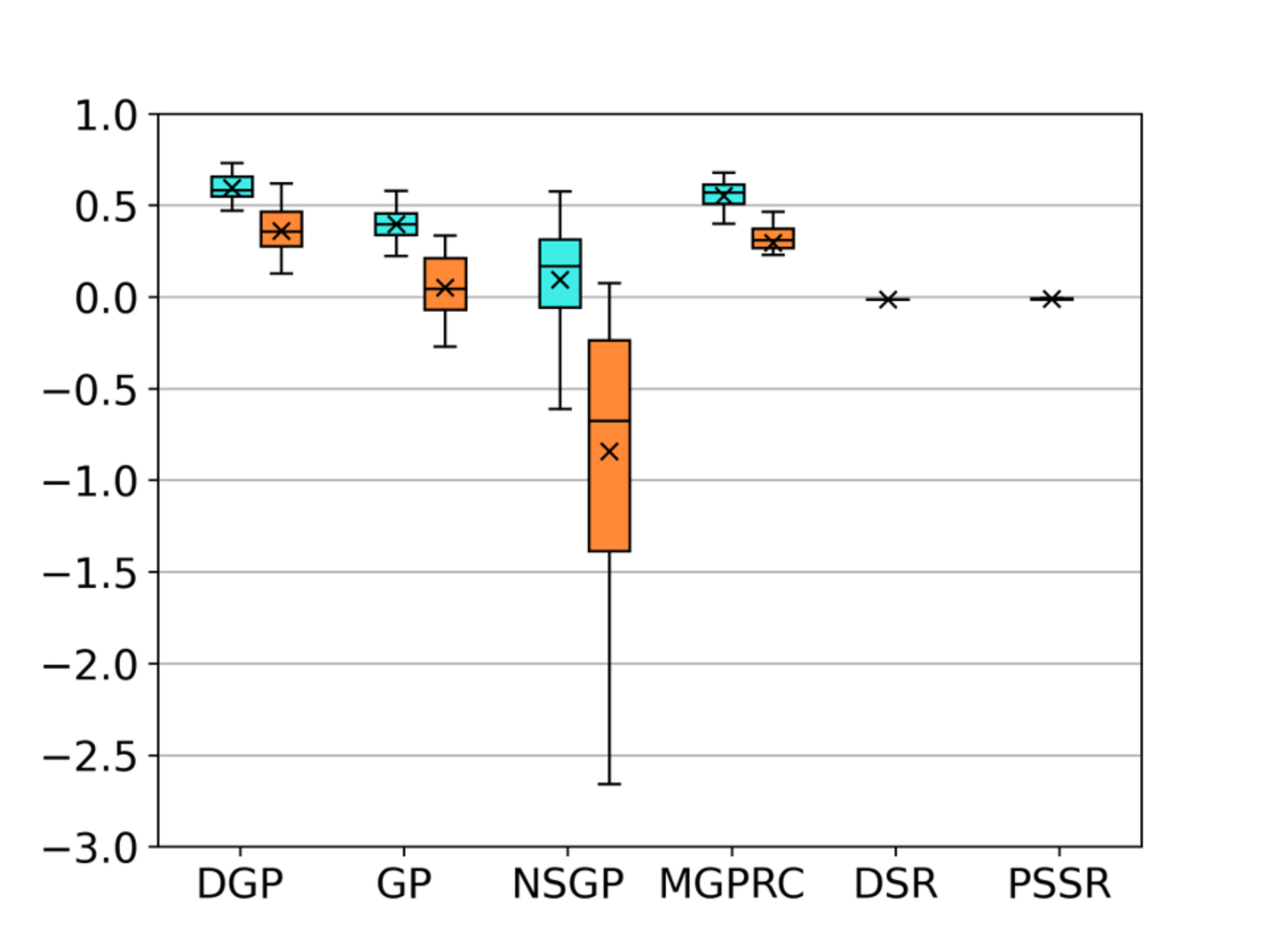}\label{fig_DLBCL_result}}
		\hfil
		\caption{Box plots on the training and test performance measured by $R^{2}$ of the found mathematical expressions by the proposed DGP method against peer competitors on the benchmark datasets. Blue and orange boxes represent the training dataset and the test dataset, respectively.}
		\label{realworld_results}
	\end{center}
\end{figure*}

The averages and standard deviations of the test $R^2$ scores on the real-world datasets are presented in Table~\ref{test r2}. The distribution of $R^{2}$ scores on the training datasets and test datasets are shown in Fig.~\ref{realworld_results}. In the figure, each method has two boxes on each dataset. Blue and orange boxes are for the training dataset and the test dataset, respectively. Results of the statistical significance tests of $R^{2}$ scores between DGP and other baseline methods are presented in Table~\ref{significant_tests}. In the table, ``$-$" indicates DGP performs significantly better than the compared method, ``$+$" indicates DGP is significantly worse, and ``$=$" means there is no significant difference. We discuss the training performance and the generalization performance in Subsections~\ref{sucsec:train perform} and~\ref{subsec:test perform}. The program size of the final expressions is compared and analyzed in Subsection~\ref{subsec:program size}.

\begin{table*}[!htbp]
	\centering
	\caption{Test $R^2$ score of DGP and other methods on the eight real-world benchmarks over 30 independent runs}
	\renewcommand{\arraystretch}{1.25} 
	\setlength\tabcolsep{3pt} 
	\begin{tabularx}{\textwidth}{ccccccccc}
		\hline
		Datasets                                                     &Satellite\_image &Fri\_c4\_50   &Fri\_c0\_50   &Fri\_c1\_50   &Fri\_c4\_100  &GeoOriMusic   &Tecator         &DLBCL \\     
		Num of Features(Dimensions) &36 &50 &50 &50 &100   &117 &124 &7,400 \\
		\hline
		\hline
		GP (Fortin et al.,2012)~\cite{Fortin.2012}                    &0.34$^{\pm0.203}$           &0.57$^{\pm0.249}$         &0.58$^{\pm0.164}$          &0.45$^{\pm0.283}$         &0.32$^{\pm0.281}$         & 0.54$^{\pm0.044}$        & 0.87$^{\pm0.043}$             &0.06$^{\pm0.177}$  \\  
		NSGP (Virgolin et al.,2020)~\cite{Virgolin.2020}              &0.45$^{\pm0.083}$           &0.55$^{\pm0.434}$         &0.69$^{\pm0.219}$          &0.58$^{\pm0.307}$         &0.24$^{\pm0.586}$         & 0.51$^{\pm0.240}$        & 0.82$^{\pm0.174}$             &-0.84$^{\pm0.706}$   \\
		MGPRC (Chen et al.,2022)~\cite{Chen.2022}                     &0.61$^{\pm0.101}$           &0.47$^{\pm0.166}$         &0.62$^{\pm0.104}$          &0.66$^{\pm0.087}$         &0.43$^{\pm0.156}$         & 0.48$^{\pm0.109}$        & 0.87$^{\pm0.067}$             &0.30$^{\pm0.131}$     \\
		DSR (Peterson et al.,2021)~\cite{Petersen.2021}               &-0.39$^{\pm0.831}$           &-2.11$^{\pm1.479}$         &-0.46$^{\pm0.625}$          &-1.13$^{\pm1.644}$         &-3.13$^{\pm0.767}$         &0.12$^{\pm0.245}$        & -0.39$^{\pm1.031}$            &-349.84$^{\pm0.000}$  \\
		PSSR (Mundhenk et al.,2021)~\cite{Mundhenk.2021}              &-0.10$^{\pm0.433}$           &-0.36$^{\pm0.931}$         &0.13$^{\pm0.518}$          &0.00$^{\pm0.178}$         &-1.19$^{\pm1.149}$         & 0.27$^{\pm0.243}$        & 0.61$^{\pm0.318}$             &-58.68$^{\pm115.657}$  \\ \hline
		
		\hline
		DGP (Ours)                                                  &\textbf{0.63$^{\pm0.118}$}   &\textbf{0.65$^{\pm0.144}$}&\textbf{0.72$^{\pm0.069}$} &\textbf{0.67$^{\pm0.087}$}&\textbf{0.58$^{\pm0.147}$}&\textbf{0.62$^{\pm0.066}$}&\textbf{0.88$^{\pm0.017}$}    &\textbf{0.32$^{\pm0.127}$}  \\
				                                                						            		          		              		       		                                            
		\hline
	\end{tabularx}
	\label{test r2}
\end{table*}

\begin{table}[!htbp]
	\centering
	\caption{Results of statistical significance tests}
	\renewcommand{\arraystretch}{1.2} 
	\begin{tabularx}{0.5\textwidth}{l|X|X|X|X|X}
		\hline
		\multirow{2}{*}{Datasets} &  \multicolumn{5}{c}{DGP(training, test)vs.} \\ 
		\cline{2-6}
		~                         & GP        & NSGP       & MGPRC      & DSR      & PSSR \\
		\hline
		\hline
		Satellite\_image          & $(-, -)$  & $(-, -)$   & $(=, =)$   & $(-, -)$ & $(-, -)$ \\
		Fri\_c4\_50               & $(=, =)$  & $(=, =)$   & $(-, -)$   & $(-, -)$ & $(-, -)$ \\
		Fri\_c0\_50               & $(-, -)$   & $(+, =)$  & $(-, -)$   & $(-, -)$ & $(-, -)$ \\
		Fri\_c1\_50               & $(-, -)$  & $(=, =)$   & $(=, =)$   & $(-, -)$ & $(-, -)$ \\
		Fri\_c4\_100              & $(-, -)$  & $(-, -)$   & $(-, -)$   & $(-, -)$ & $(-, -)$ \\
		GeoOriMusic               & $(=, =)$  & $(=, -)$   & $(-, -)$   & $(-, -)$ & $(-, -)$ \\
		Tecator                   & $(-, =)$  & $(=, =)$   & $(=, =)$   & $(-, -)$ & $(-, -)$ \\
		DLBCL                     & $(-, -)$  & $(-, -)$   & $(=, =)$   & $(-, -)$ & $(-, -)$ \\
		\hline
	\end{tabularx}
	\label{significant_tests}
\end{table}

\subsubsection{Training Performance}\label{sucsec:train perform}
The blue box plots in Fig.~\ref{realworld_results} show the training performance of the expressions found by the different methods. Overall, the training performance of the GP-based methods is better than that of NN-based methods. This is partly due to the global search capability of GP, which enables it to find reasonable solutions quickly in a large search space. However, the NN-based methods, including DSR and PSSR, have better stability, as reflected by the tighter distribution of their $R^{2}$ results on most of the datasets. These results indicate that GP-based methods are more susceptible to stochasticity, meaning they may need to be run several times in practice to get the best results. 

As shown in Table~\ref{significant_tests}, the proposed DGP method has better training performance in most cases on the training datasets than its competitors. Specifically, the $R^2$ scores of DGP is significantly higher than that of GP on six of the eight datasets (and not significantly different on the remaining two). This indicates that the proposed new gradient-based GP structure optimization method can effectively improve the training performance on high-dimensional datasets. Furthermore, DGP is also comparable to state-of-the-art GP-based SR methods including NSGP and MGPRC. In most cases, DGP has better training performance than NSGP and MGPRC, except on Fri\_c0\_50 when compared with NSGP. Finally, compared to the NN-based methods, it is obvious that the training performance of DGP is significantly better than DSR and PSSR on the real-world datasets. Through the above analysis, it can be concluded that DGP has better training performance in most cases compared with other SR methods. 

\subsubsection{Generalization Performance}\label{subsec:test perform}
As is shown in Fig.~\ref{realworld_results}, the generalization performance on the test datasets is only slightly worse than the training performance for GP-based SR methods. However, for the two NN-based methods, the generalization performance is much worse than the training performance. The mean and median of $R^{2}$ results on the test datasets are much lower, and the distributions are much bigger, too. The generalization performance of NN-based methods are even worse when tested on the extremely high-dimensional dataset DLBCL, with $R^{2}$ results below the box plots of DGP in Fig.~\ref{realworld_results}. These results show that the GP-based methods have better generalization performance than NN-based methods on high-dimensional real-world datasets under the same number of evaluations.

We now compare the differences in the generalization performance of the six methods in detail. As is shown in Table~\ref{test r2}, the mean $R^{2}$ of DGP over 30 independent runs is higher than the baselines on all the eight real-world benchmarks. The results of the significance tests in Table~\ref{significant_tests} show that the $R^{2}$ score of DGP is significantly better than DSR and PSSR on all eight datasets. Compared with the GP-based SR methods, the performance is significantly better on six of the eight datasets compared to GP, and five of the eight datasets compared to NSGP and MGPRC. Considering that the generalization performance on the test data is a more important metric, we can conclude that DGP generally achieves the best performance on the eight real-world datasets.

We have two perspectives as to why our proposed DGP method outperforms the competitors. First, the gradient descent used in DGP allows it to keep continuously searching for better solutions in a relatively small search space, improving search efficiency. Second, from another point of view, the diversifier helps by providing fresh new samples to escape local optima. This is because the gradient-based discrete distributions often concentrate their probability mass on a relatively small portion of the search space, resulting in premature convergence to locally optimal solutions. However, the genetic operations in the diversifier generate new individuals which may fall well outside the concentrated region of the search space. Combining the gradient-based optimizer and the diversifier, DGP is able to regress better solutions more efficiently.

\subsubsection{Program Size Analysis}\label{subsec:program size}
The mean and standard deviation of the size of programs found by the six methods are shown in Table~\ref{program size}. The results of statistical significant tests on the program size are listed in Table~\ref{size_significant_tests}. The results show that NN-based SR methods, especially DSR, can search for more compact models with a much smaller number of nodes than GP-based methods. However, given that the generalization performance of NN-based methods on these high-dimensional datasets is much worse than that of GP-based methods, it can be inferred that the models found by DSR are too simple. The number of nodes in individuals evolved by the standard GP method is the largest on all eight datasets and far exceeds that of other methods, likely due to bloating. The three methods that specifically control bloating -- DGP, NSGP, and MGPRC -- evolve individuals with many fewer nodes than GP. The proposed DGP is more effective than NSGP and MGPRC in controlling bloating, with fewer nodes in six of the eight datasets. The length of the symbolic expressions (number of nodes) is an important measure of interpretability, and by this measure, DGP finds more interpretable solutions.

\begin{table}[!htbp]
	\centering
	\caption{Program size of DGP and competitors on the real-world benchmarks}
	\renewcommand{\arraystretch}{1.25} 
	\begin{tabularx}{0.5\textwidth}{p{1.7cm}|p{0.8cm}|p{0.85cm}|p{0.8cm}|p{0.8cm}|p{0.6cm}|p{0.5cm}}
		\hline
		Benchmark &  \multicolumn{6}{c}{Size (\#Node of the best models)} \\ 
		\cline{2-7}
		          & DGP & GP  & NSGP & MGPRC & DSR  & PSSR \\
		\hline
		\hline
		Satellite\_image & 61$^{\pm28.9}$& 136$^{\pm63.2}$ & 65$^{\pm19.7}$ & 64$^{\pm21.5}$ & 10$^{\pm2.3}$ & 40$^{\pm42.2}$  \\
		Fri\_c4\_50      & 37$^{\pm12.0}$& 70$^{\pm24.8}$  & 72$^{\pm17.8}$ & 42$^{\pm15.6}$ & 8$^{\pm2.8}$  & 16$^{\pm16.5}$  \\
		Fri\_c0\_50      & 26$^{\pm14.1}$& 76$^{\pm39.9}$  & 86$^{\pm11.7}$ & 32$^{\pm8.9}$  & 8$^{\pm2.1}$  & 17$^{\pm12.7}$  \\
		Fri\_c1\_50      & 34$^{\pm15.5}$ &65$^{\pm40.1}$  & 80$^{\pm19.4}$ & 34$^{\pm12.0}$& 7$^{\pm2.7}$  &15$^{\pm9.2}$ \\
		Fri\_c4\_100     & 29$^{\pm14.8}$ & 71$^{\pm43.5}$ & 78$^{\pm13.8}$ & 44$^{\pm17.1}$ & 8$^{\pm1.4}$   & 20$^{\pm22.1}$ \\
		GeoOriMusic      & 39$^{\pm25.0}$ & 13$^{\pm27.8}$ & 75$^{\pm17.3}$ & 10$^{\pm12.9}$ & 8$^{\pm1.6}$   &16$^{\pm7.9}$ \\
		Tecator          &39$^{\pm21.8}$ &95$^{\pm43.8}$ &53$^{\pm18.1}$ &65$^{\pm21.3}$ &8$^{\pm2.0}$ &28$^{\pm23.9}$\\
		DLBCL            &47$^{\pm21.6}$ &36$^{\pm16.4}$ &76$^{\pm18.6}$ &33$^{\pm13.6}$ &7$^{\pm0.0}$ &5$^{\pm0.4}$\\
		\hline
	\end{tabularx}
	\label{program size}
\end{table}

\begin{table}[!htbp]
	\centering
	\caption{Results of statistical significance tests on the program size}
	\renewcommand{\arraystretch}{1.2} 
	\begin{tabularx}{0.5\textwidth}{l|X|X|X|X|X}
		\hline
		\multirow{2}{*}{Benchmark} &  \multicolumn{5}{c}{DGP vs.} \\ 
		\cline{2-6}
		~                         & GP        & NSGP       & MGPRC      & DSR      & PSSR \\
		\hline
		\hline
		Satellite\_image          & -  & =  & =  & + & = \\
		Fri\_c4\_50               & - & -   & -  & +& + \\
		Fri\_c0\_50               & -  & -  & =  & + & = \\
		Fri\_c1\_50               & - & -  & =   & + & +\\
		Fri\_c4\_100              & - & -  & -  & + & =\\
		GeoOriMusic               & +  & -   & +   & + & + \\
		Tecator                   & -  &- & -   & + & = \\
		DLBCL                     & =  & -   & +   & + & + \\
		\hline
	\end{tabularx}
	\label{size_significant_tests}
\end{table}

\begin{figure*}[htp]
	\begin{center}
		\subfloat[S1]{\includegraphics[width=0.65\columnwidth]{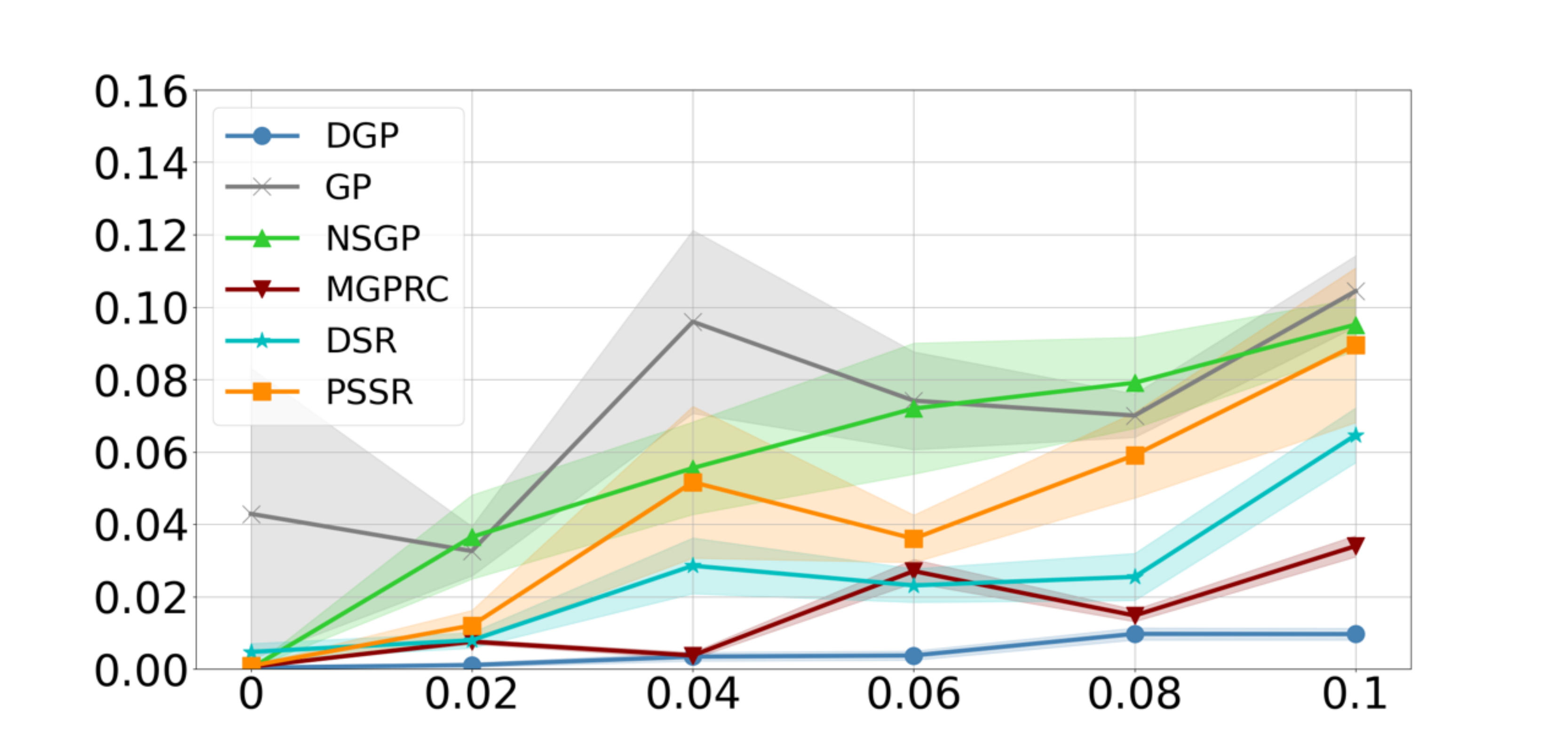}%
			\label{fig_S1_result}}
		\hfil
		\subfloat[S2]{\includegraphics[width=0.65\columnwidth]{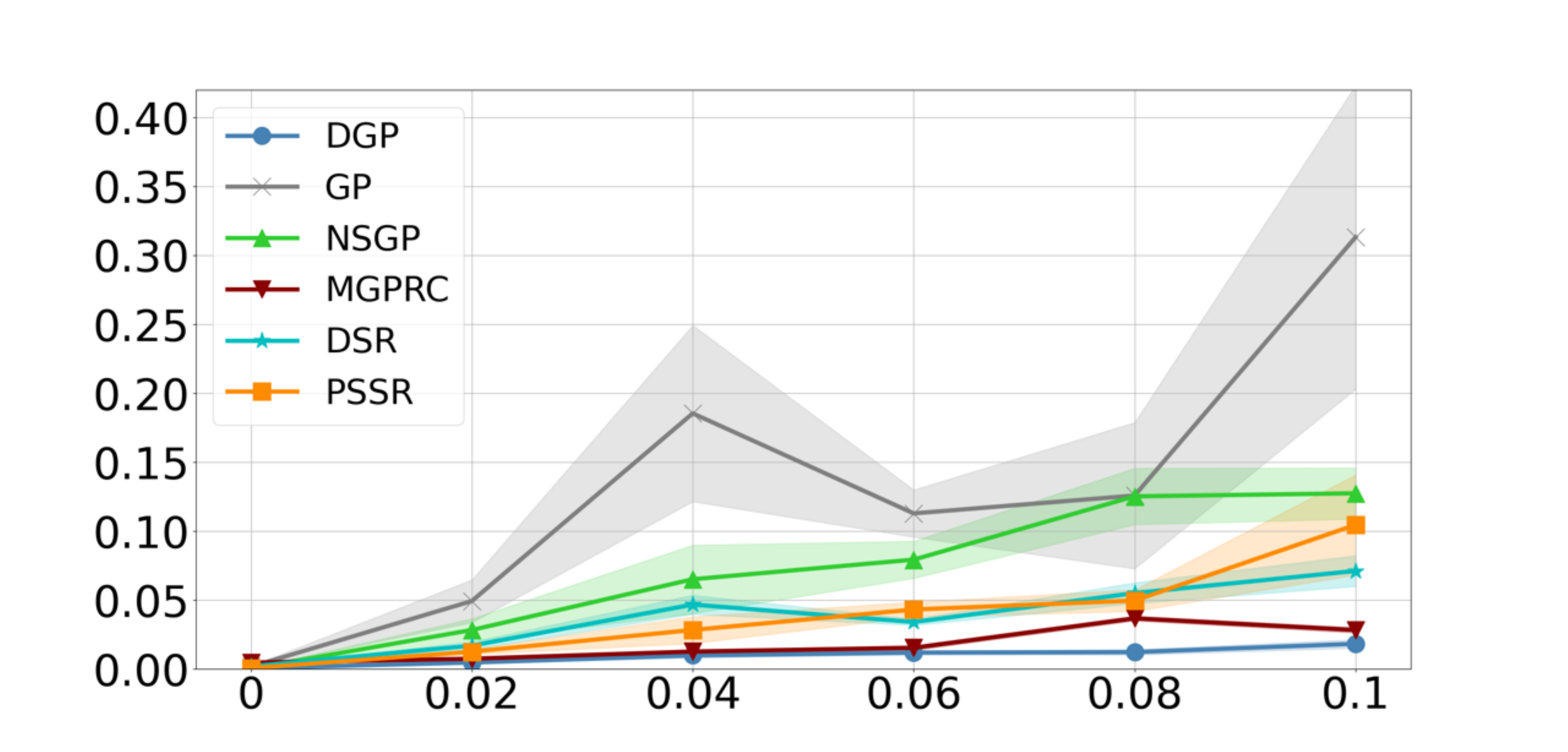}%
			\label{fig_S2_result}}
		\hfil
		\subfloat[S3]{\includegraphics[width=0.65\columnwidth]{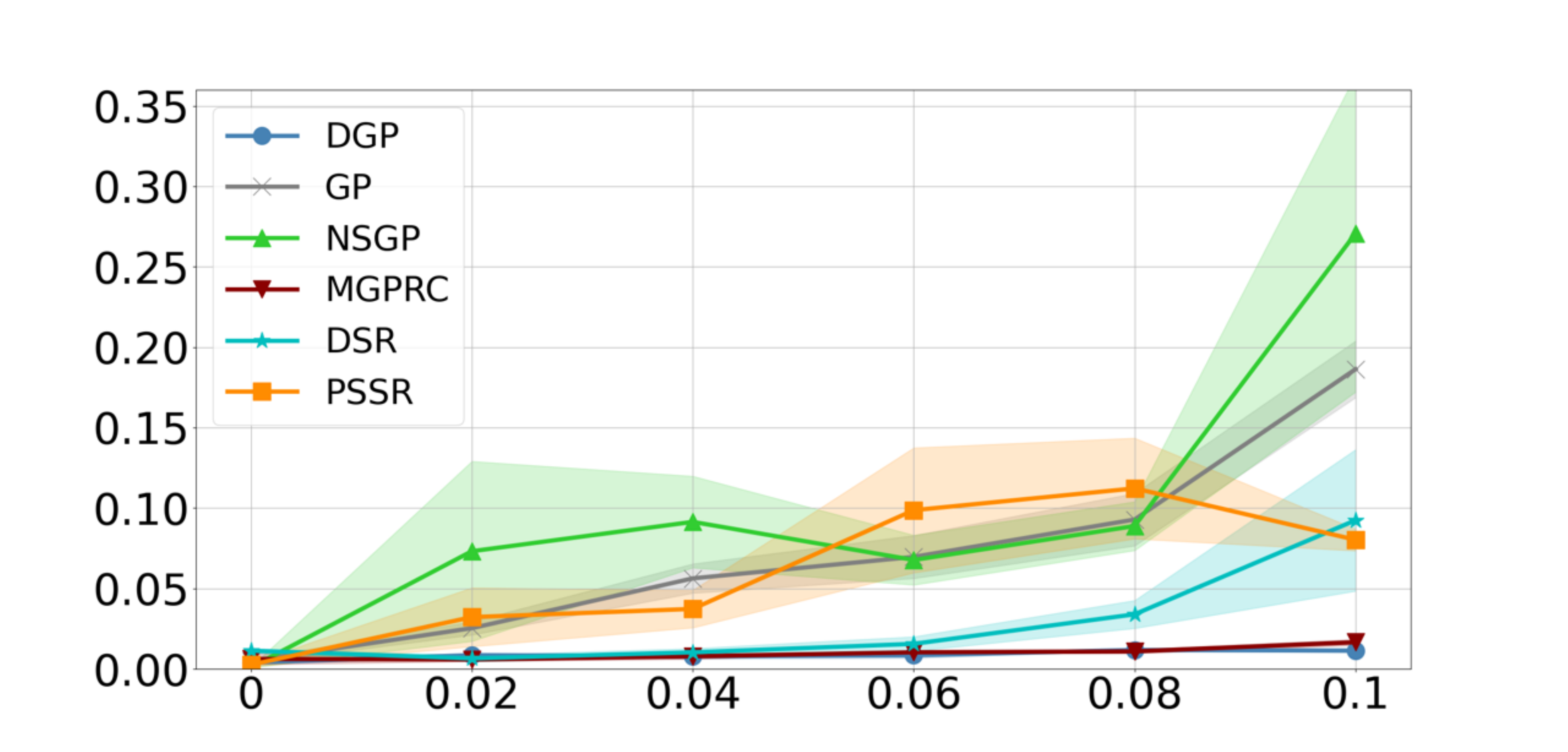}%
			\label{fig_S3_result}}
		\hfil
		\subfloat[S4]{\includegraphics[width=0.65\columnwidth]{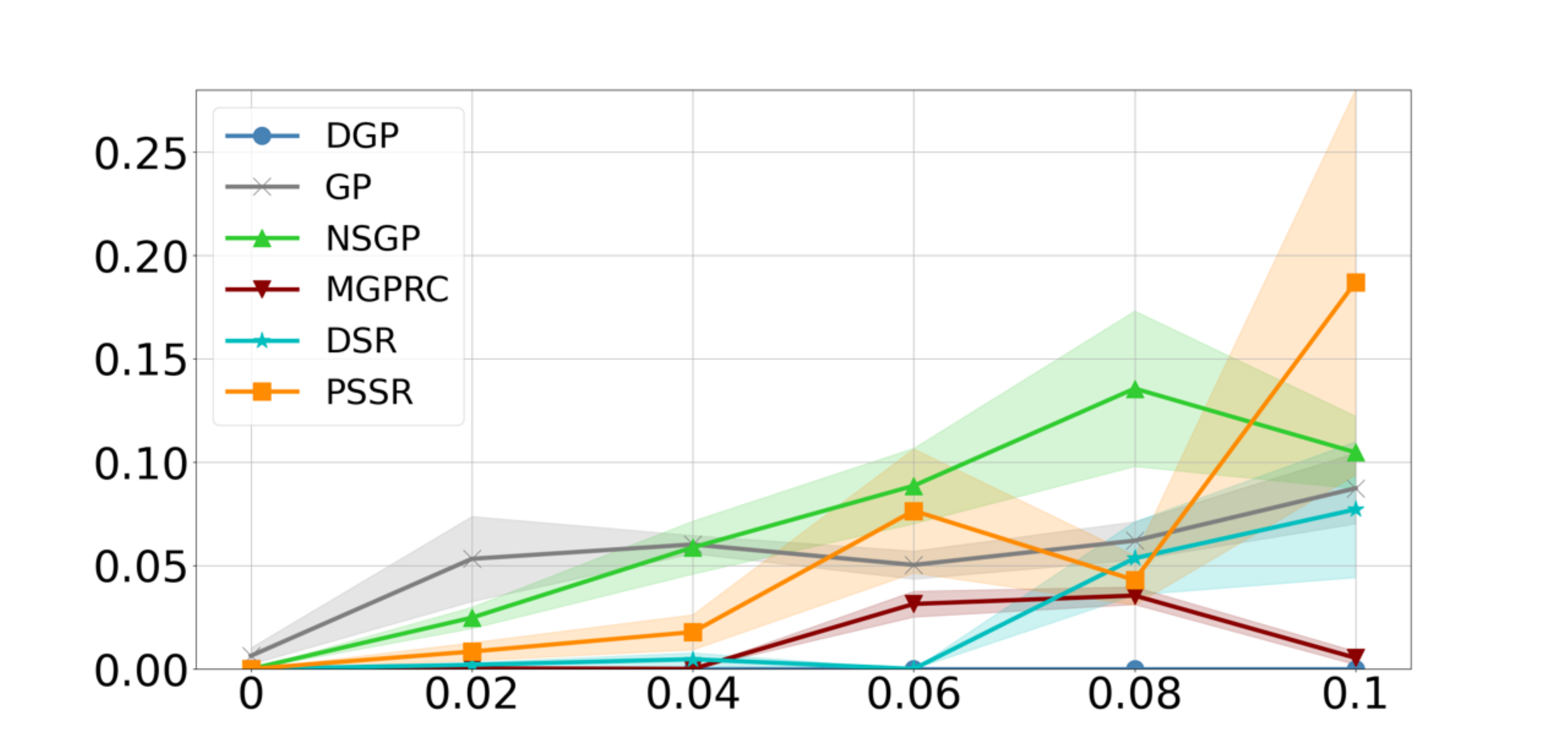}%
			\label{fig_S4_result}}
		\hfil
		\subfloat[S5]{\includegraphics[width=0.65\columnwidth]{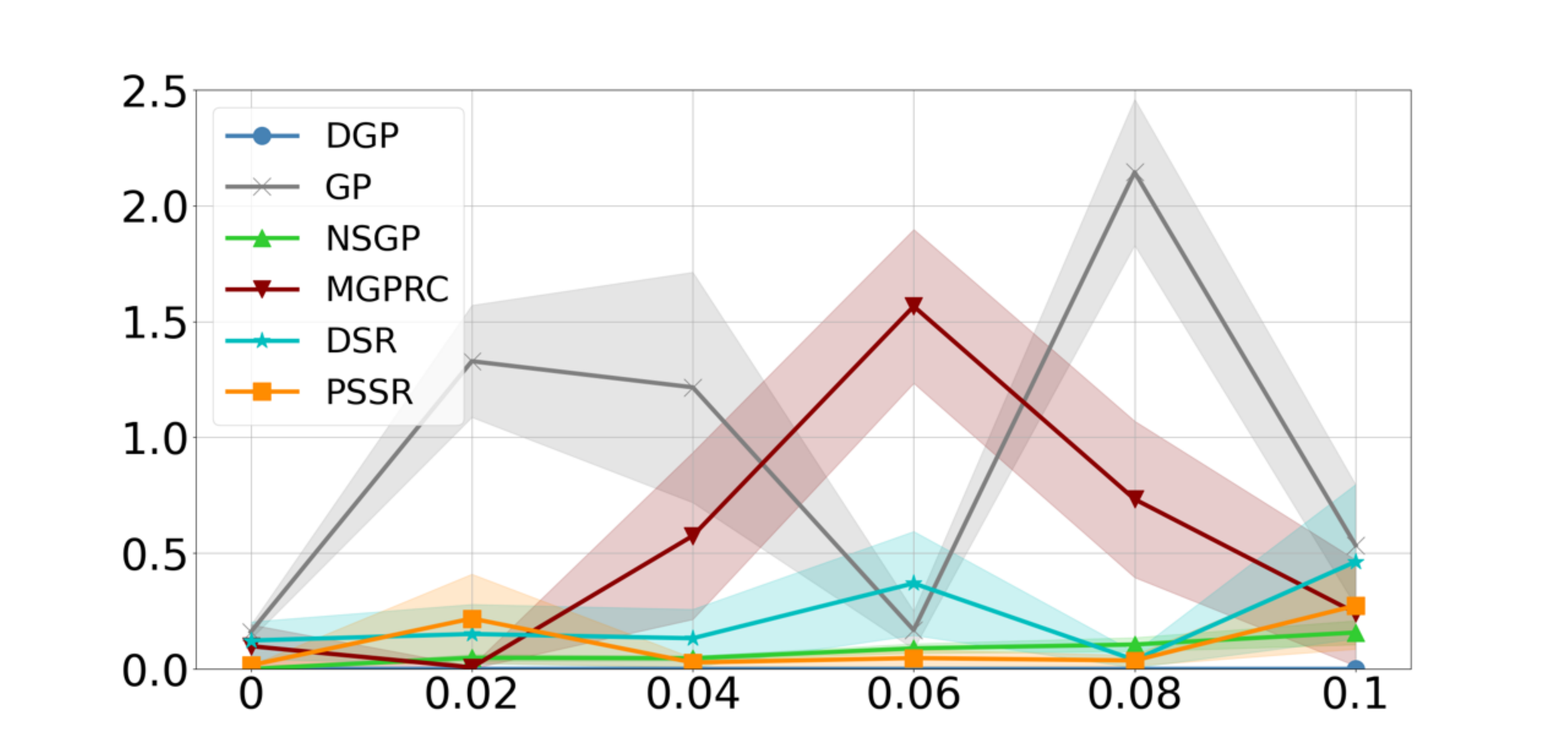}%
			\label{fig_S5_result}}
		\hfil
		\subfloat[S6]{\includegraphics[width=0.65\columnwidth]{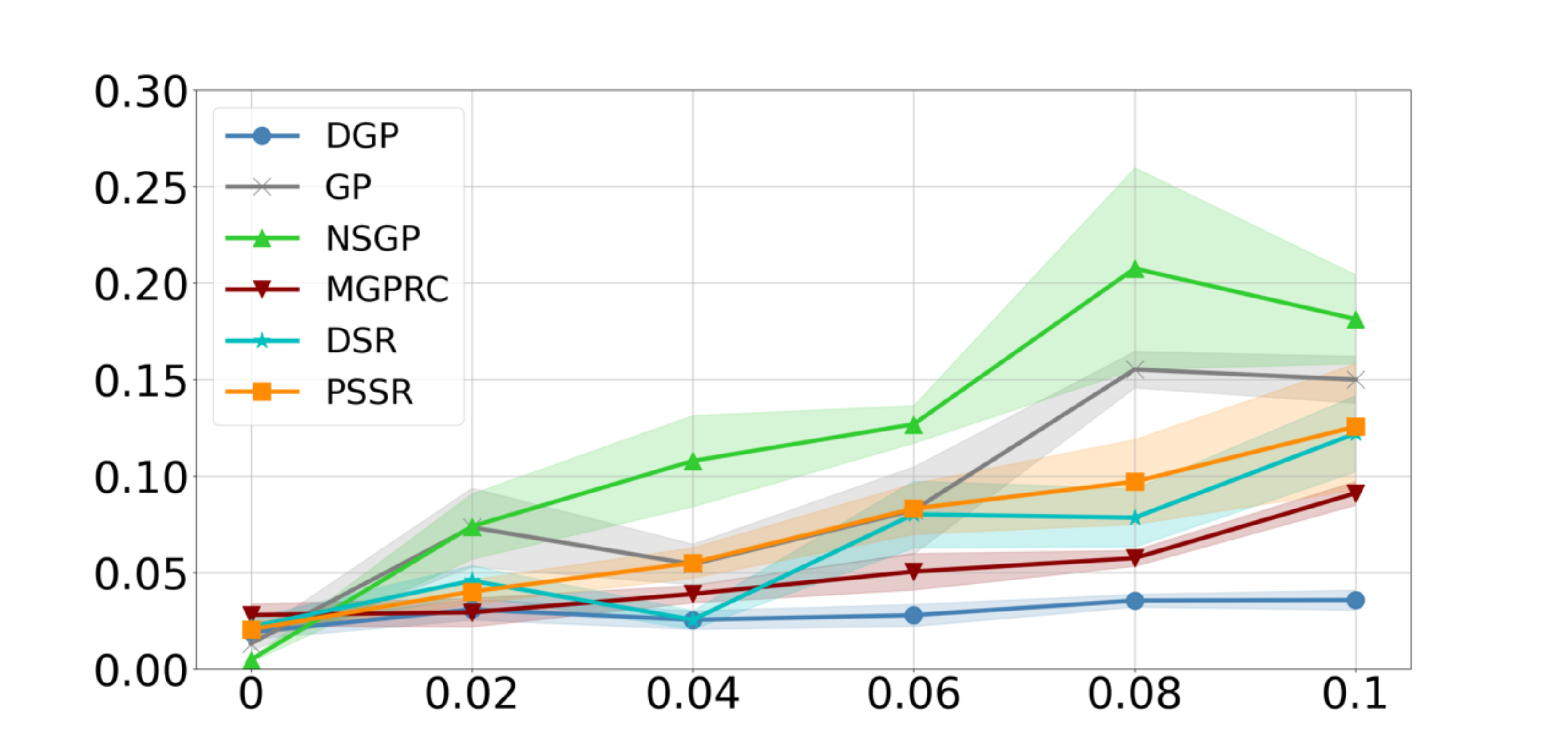}%
			\label{fig_S6_result}}
		\hfil
		
		\caption{The average test RMSE of different methods under different levels of dataset noises across all the synthetic benchmarks.}
		\label{noise_results}
	\end{center}
\end{figure*}

\subsection{Results on the Synthetic Benchmarks} \label{synthetic subsection}
In this section, we first analyse the overall recovery results of the methods on the synthetic benchmarks in Subsection~\ref{subsec:synthetic_overall}. We then analyse the program size of the expressions found by different methods in Subsection~\ref{subsec:size}. Finally, we compare the methods' robustness by adding various noise levels to the data and analysing the results in Subsection~\ref{subsec:synthetic_noise}.

\subsubsection{Recovery Results}\label{subsec:synthetic_overall}

\begin{table}[!htbp]
	\centering
	\caption{Recovery rates of DGP and competitors on the synthetic benchmarks}
	\renewcommand{\arraystretch}{1.2} 
	\begin{tabularx}{0.5\textwidth}{l|X|X|X|l|X|X}
		\hline
		 Benchmark &  \multicolumn{6}{c}{Recovery rate (\%)} \\ 
		\cline{2-7}
		          & DGP & GP  & NSGP & MGPRC & DSR  & PSSR \\
		\hline
		\hline
		S1                       & 90  & 30   & 60   & 40    & 0    & 80 \\
		S2                       & 80  & 0    & 10   & 0     & 0    & 80 \\
		S3                       & 50  & 0    & 0    & 40    & 0    &50 \\
		S4                       & 100 &50    & 100  & 100   & 100  &100 \\
		S5                       & 100 & 0    & 0    & 90    & 70   & 90 \\
		S6                       & 20  & 0    & 0    & 0     & 20   &10 \\
		\hline
		Average                 & 73.3 &13.3 &28.3 &45.0 &31.7 &68.3\\
		\hline
	\end{tabularx}
	\label{recovery rate}
\end{table}

The overall recovery results are shown in Table~\ref{recovery rate}. It can be seen from Table~\ref{recovery rate} that the mean recovery rate of the proposed DGP far exceeds that of four of the five other methods on the benchmarks, which indicates that DGP can recover accurate mathematical expressions from the data effectively. Compared to the final method (PSSR), it is better than PSSR on three benchmarks and performs equally on the remaining three. The recovery rates of traditional GP are lower than other methods and ineffective in recovering exact expressions. Traditional GP has a tendency to generate increasingly complex symbolic trees (due to the canonical evolutionary approach); thus the search process of GP often reaches a dead end. Our proposed method instead adds a certain direction to the search process, and our introduction of shrinking effectively limits the bloating of symbolic trees, meaning it efficiently searches for valid expressions.

\subsubsection{Robustness Analysis}\label{subsec:synthetic_noise}
According to the experiments in~\cite{Petersen.2021}, the search difficulty of true symbolic expressions varies with the noise level of the target variable, so it is important to demonstrate DGP's robustness to noise. We evaluate the proposed DGP on noisy data by adding independent Gaussian noise to the target variable, with mean zero and standard deviation proportional to the root-mean-square of the target variable in the training data. We found that most of the methods could not recover the exact symbolic expressions from data at a high noise level (i.e., $> 0.06$), meaning that the recovery rates are mostly zero and do not differentiate the effectiveness of the methods. Thus, we instead have reported the test RMSE (Fig.~\ref{noise_results}) of the different methods on various noise levels from 0 (noiseless) to $10^{-1}$.

As is shown in Fig.~\ref{noise_results}, the performance of different methods varies substantially across different noise levels. For the NN-based SR methods, PSSR (which is generally seen as an improvement of DSR), has a better average recovery rate on noiseless benchmarks than DSR. However, our experiments show that the robustness of PSSR is worse than DSR: the test RMSE of PSSR is higher than DSR in four of the six benchmarks. As for GP-based SR methods, the test RMSE of GP and NSGP clearly increases with the increasing levels of noise, showing they are not robust. The robustness of MGPRC is quite promising. It can be seen from Fig.~\ref{noise_results} that the test RMSE of MGPRC increase slowly with the increasing noise level and is second place on all the synthetic benchmarks except for S5. Finally, the robustness of our proposed DGP is convincing. Although the noise in the training data has some influence, the performance attenuation of DGP is relatively minor. At the highest noise level ($10^{-1}$), DGP achieves the lowest test RMSE of the six SR methods on all the synthetic benchmarks.

These results make DGP a very promising method for SR: it is both highly accurate and robust to noise across a range of datasets.

\subsubsection{Program Size Analysis}\label{subsec:size}
In this section, we compare the program size of the symbolic expressions found by the different methods, where methods were able to successfully recover the expression from the data (i.e., where recovery rate was greater than 0\%). This is due to the fact that it is meaningless to compare the size of expressions if the correct one cannot be found. The results are presented in Table~\ref{program size_2}.
\begin{table}[!htbp]
	\centering
	\caption{Program size of DGP and competitors on the synthetic benchmarks}
	\renewcommand{\arraystretch}{1.25} 
	\begin{tabularx}{0.5\textwidth}{p{1.25cm}|p{0.8cm}|p{0.9cm}|p{0.65cm}|p{0.8cm}|p{0.65cm}|p{1.0cm}}
		\hline
		Benchmark&  \multicolumn{6}{c}{Size(\#Node of the best models)} \\ 
		\cline{2-7}
		                 & DGP & GP  & NSGP & MGPRC & DSR  & PSSR \\
		\hline
		\hline
		S1                       & 19$^{\pm5.4}$ & 108$^{\pm7.1}$ & 14$^{\pm1.9}$ & 22$^{\pm3.8}$  & -  & 19$^{\pm4.2}$ \\
		S2                       & 34$^{\pm10.4}$& -    & 42$^{\pm0.0}$   & -     & -    & 26$^{\pm4.4}$ \\
		S3                       & 22$^{\pm3.1}$  & -    & -    & 41$^{\pm18.5}$    & -    &18$^{\pm1.6}$ \\
		S4                       & 9$^{\pm2.0}$ &123$^{\pm51.0}$ & 12$^{\pm9.7}$ & 23$^{\pm8.1}$ & 7$^{\pm0.0}$  &11$^{\pm4.9}$ \\
		S5                       & 14$^{\pm4.7}$ & -    & -    & 32$^{\pm10.5}$  & 22$^{\pm9.6}$ & 21$^{\pm9.3}$ \\
		S6                       & 19$^{\pm5.5}$  & -    & -    & -     & 13$^{\pm1.5}$   &29$^{\pm0.0}$ \\
		\hline
	\end{tabularx}
	\label{program size_2}
\end{table}

Table~\ref{program size_2} shows that the overall program size of NN-based methods are smaller than the GP-based methods, except for DGP. NN-based methods are able to find more compact expressions as they use an RNN to gradually generate symbolic trees starting from a simple structure, which effectively controls the program size of expressions. Canonical GP again has the largest programs of all the GP methods, indicating that there are many redundant structures (introns) in the expressions it finds. The other two GP-based baselines (NSGP and MGPRC) find more compact models than the canonical GP on the synthetic benchmarks. The program size of the expressions found by DGP is also much smaller than the canonical GP and can rival with the NN-based SR methods. We conclude that the symbolic expressions recovered by the proposed DGP method can achieve high accuracy while being concise.

\section{Conclusions and Future work}\label{conclusion}
The objective of this article was to propose an effective GP method for SR, which can avoid the ineffectiveness and bloating of traditional GP caused by the stochastic nature of its evolutionary-based search approach. We have successfully achieved this goal by proposing a new differentiable genetic programming (DGP) method, which uses a gradient descent approach to optimize the structure of GP. For the gradient-based optimization, a new representation named the differentiable symbolic tree was proposed that relaxes the discrete structure into a continuous space. In addition, we also designed a loss function for SR, proposed a unique hybrid forward propagation approach, and formulated the backward gradient calculation. With this design, the proposed DGP can search for the best structure more efficiently, making it an effective method to deal with high-dimensional symbolic regression problems compared to the state-of-the-art. The proposed method was compared to benchmark methods that included both GP-based and NN-based SR approaches by testing on both real-world and synthetic benchmark datasets. The experiment results showed that the training and generalization performance of DGP outperforms almost all the other GP-based and NN-based SR methods, and that DGP produces a relatively small solution size compared to other GP-based methods. This, in conjunction with further robustness testing, demonstrated that the proposed DGP is an effective tool for solving SR, especially for high-dimensional real-world problems that are difficult to solve with previous GP-based methods.

We note that the proposed gradient-based optimization approach for GP is not only applicable for SR tasks but also for other GP-based machine learning problems, which should be explored in future work. In addition, our future work will explore simultaneously optimizing constants and tree structures during the training process to improve the performance of the proposed method on SR problems with constants. Finally, our proposed method makes use of the final output of the expression to compute the loss even in the early stage of the search -- this potentially may lead to the search falling into a local optimum. Thus, investigating how to combine the search for partial expressions with the search for overall expressions is also a potential research direction.

\ifCLASSOPTIONcaptionsoff
  \newpage
\fi

\bibliographystyle{IEEEtran}
\bibliography{DiSR}

\end{document}